\documentclass[10pt]{article}

\usepackage{microtype}
\usepackage{graphicx}
\usepackage{subfigure}
\usepackage{booktabs} 
\usepackage{makecell}

\usepackage{hyperref}

\usepackage{amsmath}
\usepackage{amssymb}
\usepackage{mathtools}
\usepackage{amsthm}
\usepackage{natbib}

\usepackage[capitalize,noabbrev]{cleveref}

\usepackage{fancyhdr}
\usepackage{xcolor} 
\usepackage{algorithm}
\usepackage{algorithmic}
\usepackage{natbib}
\usepackage{eso-pic} 
\usepackage{forloop}
\usepackage{url}
\usepackage[T1]{fontenc}
\usepackage[bottom]{footmisc}

\usepackage[letterpaper]{geometry}  
\usepackage{times}     

\usepackage{titlesec}
\titlespacing*{\section}{0pt}{*2.0}{*1.0}  
\titlespacing*{\subsection}{0pt}{*1.5}{*1.0}
\titlespacing*{\subsubsection}{0pt}{*1.0}{*0.5}

\theoremstyle{plain}
\newtheorem{theorem}{Theorem}[section]
\newtheorem{proposition}[theorem]{Proposition}
\newtheorem{lemma}[theorem]{Lemma}

\theoremstyle{definition}

\newtheorem{assumption}[theorem]{Assumption}
\theoremstyle{remark}
\newtheorem{remark}[theorem]{Remark}

\newcommand{\yrcite}[1]{\citeyearpar{#1}}
\setcitestyle{authoryear,round,citesep={;},aysep={,},yysep={;}}

\renewenvironment{abstract}
{	\centerline{\large\bf Abstract}
	\vspace{-0.12in}\begin{quote}}
	{\par\end{quote}\vskip 0.12in}

\title{Single Point-Based Distributed Zeroth-Order Optimization with a Non-Convex Stochastic Objective Function}

\author{
	Elissa Mhanna* \and Mohamad Assaad*
}

\date{}

\begin{document}
	\twocolumn[
	\maketitle
		]
	\begin{abstract}
		Zero-order (ZO) optimization is a powerful tool for dealing with realistic constraints. On the other hand, the gradient-tracking (GT) technique proved to be an efficient method for distributed optimization aiming to achieve consensus. However, it is a first-order (FO) method that requires knowledge of the gradient, which is not always possible in practice.
		In this work, we introduce a zero-order distributed optimization method based on a one-point estimate of the gradient tracking technique. 
		We prove that this new technique converges with a single noisy function query at a time in the non-convex setting. We then establish a convergence rate of $O(\frac{1}{\sqrt[3]{K}})$ after a number of iterations K, which competes with that of $O(\frac{1}{\sqrt[4]{K}})$ of its centralized counterparts. Finally, a numerical example validates our theoretical results.
	\end{abstract}

\let\thefootnote\relax\footnotetext{In this version, we address an imprecision in the proof of Theorem 3.7 encountered in the original publication. We remove the expectation in the proof that was added by error.}

\footnotetext{*Université Paris-Saclay, CNRS, CentraleSupélec, Laboratoire des signaux et systèmes, 91190, Gif-sur-Yvette, France. Correspondence to: Elissa Mhanna \textless elissa.mhanna@centralesupelec.fr\textgreater,  Mohamad Assaad \textless mohamad.assaad@centralesupelec.fr\textgreater}

\section{Introduction}
\label{introduction}
We consider a network encompassing a set of agents $\mathcal{N} = \{1, 2, \ldots , n\}$, where the communication is limited to network neighbors. Each agent $i$ maintains a local objective function $f_i: \mathbb{R}^d\rightarrow\mathbb{R}$ and the aim is for the agents to collaboratively find the decision variable $x\in\mathbb{R}^d$ that minimizes the stochastic global objective function:
\begin{equation}\label{objective}
	\min_{x\in\mathbb{R}^d} \mathcal{F}(x)=  \frac{1}{n}\sum_{i=1}^{n}F_i(x) 
\end{equation}
with
\begin{equation}\label{F_i}
	F_i(x)=\mathbb{E}_{\xi\sim D_i} f_i(x,\xi),
\end{equation}
where $\xi$ denotes an i.i.d. ergodic stochastic process that follows a local distribution $D_i$. Hence, the goal is for agents to achieve consensus while the local objective functions are kept private. Further, these functions are assumed to be non-convex, and to solve this problem, agent $i$ can only query the function values of $f_i$ at exactly \textit{one point} at a time. These function queries are assumed to be noisy $\tilde{f}_{i}=f_{i}+\zeta_{i}$ with $\zeta_{i}$ some additive noise.
We note that $\zeta_{i}$ is added to differentiate between the stochastic nature of $f_i$ and the noise imposed on its querying. While the expectation of the local function might not change with this addition, its variance increases, presenting an added difficulty to the analysis.

Distributed optimization appears across many applications, in distributed data processing and control in sensor networks, parallel computing and data storing, and big-data analytics. It has many interests in machine learning applications such as data rights, privacy, and scalability \cite{DSGD, nedic-ML}. One important technique in this domain is the gradient tracking method, which is agnostic to the data distribution \cite{nc-gr-tr}, meaning there's no need for data to be i.i.d across all the devices. In addition, this technique was shown to have interesting rates. For example, it converges linearly to the optimal solution with constant step size \cite{gr-tr1, gr-tr3,shipu}. 

However, all these references assume the availability of first-order information, which is not always the case, especially when gradient computation can be impractical like when the system is constantly changing, costly when the system is too complex or too huge in dimension, or even impossible such as in black-box optimization. For example, when training a deep neural network, the loss output is an intricate function of the weights, which complicates the derivation of the gradient \cite{DNN}. In bandit settings, an adversary reveals a cost value at every period, and the player must compete with this adversary without any knowledge of the underlying cost function \cite{ref1}. In centralized and distributed settings, much work has been done on zero-order (ZO) optimization. ZO methods include substituting the gradient by one-point estimates with one functional value at a time \cite{1,ref1, CZO, M-A}, generally of the form
\begin{equation}\label{1p}
	g= \frac{d}{\gamma}f(x + \gamma z, \xi) z,
\end{equation}	
with $\gamma>0$ a small value and $z$ a random vector with symmetrical distribution.
It includes	two- and multi-points estimates \cite{ref2-ref, CZO, CZO2, ref7} having the respective forms 
\begin{equation}\label{2p}
	g=d\frac{f(x+\gamma z, \xi)- f(x-\gamma z, \xi)}{2\gamma} z
\end{equation}
\begin{equation}\label{mp}
	\text{and}\; g=\sum_{j=1}^{d}\frac{f(x+\gamma e_j, \xi)- f(x-\gamma e_j, \xi)}{2\gamma} e_j
\end{equation}
where $\{e_j\}_{j=1,\ldots,d}$ is the canonical basis and other techniques which are derivative-free \cite{ref3, bubeck}. 

While the gradient tracking technique has been extended to the ZO case \cite{ref7, M-A}, the prior utilizes a multi-point estimator of the gradient and a static objective function \textit{without} any stochasticity/noise whatsoever in the system, and the latter deals with the problem in the \textit{strongly convex} setting. Both of which do not satisfy the assumptions of our problem.

As various stochastic sources and noisy environments can be present in the model, such as distinct device data distributions, uncertainties in electronic devices, discretization errors, lossy data compression, and time-varying communication channels, static objective functions no longer serve realistic purposes. 	
Moreover, multi-point estimates assume that it is possible to observe multiple realizations of the objective function under the same system conditions, as can be seen from (\ref{2p}) and (\ref{mp}), many function queries are done for the same realization of $\xi$. This assumption does not hold when the environment is continually changing. For example, in wireless networks, when the query is an estimate is subject to a wireless channel between the transmitter and the user,  that usually changes every $1-2$ ms. Thus, by the time the next observation is queried, a stochastic process will have already impacted the new function value.

While the distributed convex setting is present in many applications, countless problems call for non-convex optimization, such as in distributed learning \cite{distr_ML} and resource allocation \cite{resource-allocation}.	

For all these reasons, we extend the gradient tracking technique to stochastic objective functions and one-point gradient estimators in the non-convex setting.

\subsection{Related Work}

\begin{table*}[t]
	\caption{A summary of convergence rates attained in related work.}
	\label{sample-table}
	\vskip 0.15in
	\begin{center}
		\begin{small}\small
			\begin{sc}
				\begin{tabular}{lccccr}
					\toprule
					Reference & \makecell{Gradient\\Estimate} & Decision & GT & Setting & \makecell{Convergence\\Rate} \\
					\midrule
					Flaxman et al. \yrcite{ref1}    & One-point & Centralized & - & Convex & $O(\frac{1}{\sqrt[4]{K}})$ \\
					Roy et al. \yrcite{CZO} & One-point & Centralized & - & Non-Convex & $O(\frac{\sqrt[3]{W_K}}{\sqrt[3]{K}})$ \\
					Balasubramanian and Ghadimi \yrcite{CZO2} & Two-point & Centralized & - & Non-Convex & $O(\frac{1}{\sqrt[4]{K}})$ \\
					Li and Assaad \yrcite{1}    & One-point & Distributed & $\times$ & Strongly Convex & $O(\frac{1}{\sqrt{K}})$ \\
					Tang et al. \yrcite{ref7}    & Two-point & Distributed & $\times$ & Non-Convex & $O(\frac{1}{\sqrt{K}}\log K)$ \\
					Pu and Nedi\'{c} \yrcite{nedic}    & Unbiased/BV & Distributed & $\surd$ & Strongly Convex & $O(\frac{1}{K})$ \\
					Tang et al. \yrcite{ref7}    & $2d$-point & Distributed & $\surd$ & Non-Convex & $O(\frac{1}{K})$ \\
					Mhanna and Assaad \yrcite{M-A}    & One-point & Distributed  & $\surd$ & Strongly Convex & $O(\frac{1}{\sqrt{K}})$ \\
					This Paper    & One-point & Distributed & $\surd$ & Non-Convex & $O(\frac{1}{\sqrt[3]{K}})$ \\
					\bottomrule
				\end{tabular}
			\end{sc}
		\end{small}
	\end{center}
	\vskip -0.1in
\end{table*}

\textbf{Centralized ZO Optimization}: 
Adversarial convex bandit problems have been long studied in the zero-order domain. Flaxman et al. introduce a one-point gradient estimator to the online gradient descent technique to compete with oblivious and adaptive adversaries, which accomplishes a regret bound of  $O(K^{\frac{3}{4}})$ \yrcite{ref1}. Agarwal et al. then extend the problem to the multi-point gradient estimate \yrcite{ref2-ref}; In the noise-free setting, this estimate has bounded and sometimes vanishing variance. Their algorithm acquires regret bounds of $O(\sqrt{K})$ and of $O(\log(K))$ for strongly convex loss functions. Nesterov and Spokoiny study the complexity with a two-point estimator in different noise-free settings \yrcite{ref4}. In a stochastic convex case and another static non-convex case, a convergence rate of $O(\frac{1}{\sqrt{K}})$ is attained.

Roy et al. present and examine noise-free stochastic algorithms for general non-convex stochastic objective functions that change with time \yrcite{CZO}. They propose nonstationary versions of regret measures and define $W_K$ as a bound on the amount of nonstationarity. $W_K$ is allowed to increase with $K$. They present a gradient descent algorithm that achieves a regret bound of $O(\sqrt{K W_K})$ with a two-point gradient estimator and $O(K^{\frac23}W_K^{\frac13})$ with a one-point estimator.

Balasubramanian and Ghadimi also propose and investigate zeroth-order algorithms for stochastic objective functions \yrcite{CZO2} in the noiseless setting. They employ structural sparsity assumptions in gradient descent to deal with high dimensions. With a two-point estimate, they acquire a rate of $O(\frac{1}{\sqrt[4]{K}})$ for non-convex optimization and $O(\frac{1}{\sqrt{K}})$ for convex one. 

\textbf{Distributed ZO Optimization}: Li and Assaad present a distributed stochastic gradient descent technique in networks with a one-point gradient estimator based on a stochastic perturbation \yrcite{1}. The communication in the network is restricted to a partial exchange of the observation of the agents' local objectives that each agent uses to estimate the global objective. This objective is subject to a stochastic process, and its solution is distributed where each agent is responsible for optimizing its own scalar variable. Their convergence rate evolves as $O(\frac{1}{\sqrt{K}})$ with the assumption of a strongly concave objective function. 

Tang et al. present two consensus-aiming stochastic multi-agent optimization techniques in the static non-convex setting \yrcite{ref7}. One technique is based on a $2$-point gradient estimator without gradient tracking, and the other is based on a noise-free $2d$-point with gradient tracking. The first achieves a rate of $O(\frac{1}{\sqrt{K}}\log K)$.

\textbf{Gradient Tracking with FO Information}: All Qu and Li \yrcite{gr-tr1}, Lorenzo and Scutari \yrcite{gr-tr2}, Nedic et al. \yrcite{gr-tr3}, Shi et al. \yrcite{gr-tr4}, Li et al. \yrcite{gr-tr5}, and Jiang et al. \yrcite{gr-tr6} present a variant of the technique with the assumption of accurate gradient information availability. 

Other references assume access to local stochastic first-order oracles where the gradient estimate is unbiased, and with a bounded variance \cite{nedic, st-gr-tr, shipu, st-gr-tr2, nc-gr-tr}. The first three references consider smooth and strongly convex local objective functions, and the convergence rate is linear when the step size is constant. When the step size is decreasing, it is that of $O(\frac{1}{K})$ \cite{nedic}. Pu and Nedi\'{c} propose a distributed stochastic gradient tracking method (DSGT) and a gossip-like stochastic gradient tracking method (GSGT) that allows the agent to wake up at each round with a certain probability \yrcite{nedic}. Xin et al. extend the technique to strongly connected graphs \yrcite{st-gr-tr}. 
Pu then extends it to directed graphs assuming noisy information exchange between agents \yrcite{shipu}.
Both Lu et al. \yrcite{st-gr-tr2} and Koloskova et al. \yrcite{nc-gr-tr} study the technique in the non-convex stochastic setting and achieve a convergence rate of $O(\frac{1}{\sqrt{K}})$ under constant step size.

While those are all good rates, comparing with them is not fair as there is naturally a gap in the convergence rate between first-order and zero-order methods. In zero-order techniques, a biased estimate of the gradient is obtained, contrary to first-order methods, where an unbiased estimate is considered. Besides, the use of a constant step size guarantees a convergence near the optimal/stationary point and not to the point itself.

\textbf{Gradient Tracking with ZO Information}: In their gradient tracking method, Tang et al. consider a non-convex static setting \yrcite{ref7}. They employ a noise-free $2d$-point gradient estimator, which has a vanishing variance, and their algorithm acquires a rate of $O(\frac{1}{K})$ with constant step size. Mhanna and Assaad then extend this technique to a one-point gradient estimate achieving a rate of $O(\frac{1}{\sqrt{K}})$, but they assume strong convexity of the objective function \yrcite{M-A}.

\subsection{Challenges and Contribution}
The first challenge of our work is that when the objective function is stochastic, the underlying stochastic law is not supposed to be known. Thus, the expectation cannot be computed. And when the observation is noisy, the algorithm's performance would be especially affected as both the variance and the norm squared of the gradient estimate would increase. Our next obstacle involves analyzing the convergence of existing algorithms in non-convex optimization problems without relying on the assumption that the functions involved are strongly convex. In this scenario, convergence analysis requires us to verify that the norm of the exact gradient approaches zero. However, due to the bias in our gradient estimate w.r.t. to the true gradient, even if the estimate approaches zero, it remains challenging to prove that the exact gradient also converges to zero. While all estimates are subject to bias, multi-point cases differ because their variance $\mathbb{E}[\|\nabla\mathcal{F}-g\|^2]$ is always bounded when there is no noise present in the optimization process and can sometimes be vanishing. However, in general, for the standard single-point estimate in (\ref{1p}), this variance cannot be bounded \cite{z-o}, meaning the estimate can deviate significantly from the true gradient. This demonstrates both technical and intuitive difficulties. Unlike convex optimization, where the characteristics of the function can guide the algorithm, non-convex optimization lacks such guidance. Additionally, we face the challenge of having access only to the expectation of our gradient estimate at a specific point while aiming to prove that the gradient at the agents' average of that point vanishes and that all agents eventually reach consensus. This further complicates the problem. Nevertheless, we successfully overcome all of these challenges in our analyses of convergence and rate.

In this paper, we focus on single-point distributed zero-order optimization, which is motivated by the fact that it's not always possible to query multiple points, especially in stochastic environments, as mentioned earlier. Our contribution can then be summarized as follow:
\begin{itemize}
	\item We extend the gradient tracking technique in the stochastic non-convex setting to the case where we don't have FO information and have to estimate the gradient with a noisy one-point query.
	\item We overcome the technical challenges of non-convexity, the stochastic impact of the environment, and biasedness of our gradient estimate.
	\item We prove the convergence of the proposed algorithm theoretically using stochastic approximation techniques.
	\item We prove a convergence rate of $O(\frac{1}{\sqrt[3]{K}})$ is attainable using this algorithm which is comparable to that of ZO non-convex methods as shown in Table \ref{sample-table}.
\end{itemize}

\subsection{Notation}
Vectors are column-shaped unless otherwise specified and $\mathbf{1}$ denotes the vector of all entries equal to $1$.  For two vectors $a$, $b$ $\in\mathbb{R}^{n}$, $\langle a,b\rangle$ is the inner product. For two matrices $A$, $B\in\mathbb{R}^{n\times d}$, $\langle A,B\rangle$ is the Frobenius inner product. $\|.\|$  is	the $2$-norm for vectors and the Frobenius norm for matrices.

\subsection{Problem Assumptions}
In this section, we introduce the assumptions necessary for convergence of our algorithm. 
\begin{assumption}\label{local_fcts}
	(the local functions) We assume the Lipschitz continuity of all local objective functions $x\longmapsto f_i(x,\xi)$,
	\begin{equation}
		\|f_i(x,\xi)-f_i(x',\xi)\|\leq L_\xi\|x-x'\|, \; \forall x,x'\in\mathbb{R}^d, \forall i\in\mathcal{N},
	\end{equation}
	where $L_\xi$ denotes the Lipschitz constant.
	Further, we assume $\mathbb{E}_{\xi} f_i (x, \xi) < \infty, \forall i \in\mathcal{N}$, to ensure the boundedness of the objective $\mathcal{F}(x)$.
\end{assumption}
\begin{assumption}\label{obj_fct}
	(the objective function) 
	Both $\nabla F_i(x)$ and $\nabla^2 F_i(x)$ exist and are continuous, and there exists a constant $\sigma_1>0$ such that
	\begin{equation*}
		\|\nabla^2 F_i(x)\|\leq\sigma_1, \;\forall x\in\mathbb{R}^d, \forall i\in\mathcal{N}.
	\end{equation*}
	Hence, we can say that the stochastic objective function $x\longmapsto \mathcal{F}(x)$ is $L$-smooth for some positive constant $L$,
	\begin{equation}\label{smooth}
		\|\nabla\mathcal{F}(x)-\nabla\mathcal{F}(x')\|\leq L\|x-x'\|, \;\forall x,x'\in\mathbb{R}^d,
	\end{equation}
	or alternatively, 
	\begin{equation}\label{smooth2}
		\mathcal{F}(x)
		\leq \mathcal{F}(x')+\langle\nabla \mathcal{F}(x'), x-x'\rangle +\frac{L}{2}\|x-x'\|^2.\\
	\end{equation}
\end{assumption}
\begin{assumption}\label{noise}
	(the additive noise) $\zeta_{i}$ is a zero-mean uncorrelated noise with bounded variance, meaning $E(\zeta_{i}) = 0$, $E(\zeta_{i}^2)=\sigma_2<\infty$, $\forall i\in\mathcal{N}$, and $E(\zeta_{i}\zeta_{j}) = 0$ if $i\neq j$.
\end{assumption}
\begin{assumption}\label{network}(the graph)
	The network is described by an undirected and connected graph, meaning communication links work in both directions, and between any two agents, we can find a path of links.  
	
	We then define the agents' coupling matrix $W = [w_{ij}] \in \mathbb{R}^{n\times n}$, where
	agents $i$ and $j$ are connected iff $w_{ij} = w_{ji} > 0$ ($w_{ij} = w_{ji} = 0$ otherwise).
	We assume $W$ is a nonnegative matrix and doubly stochastic, i.e., $W \mathbf{1} = \mathbf{1}$ and $\mathbf{1}^T W = \mathbf{1}^T$. All diagonal elements $w_{ii}$ are strictly positive.
\end{assumption}
\section{Algorithm Description}\label{sec-algo}
This section describes our proposed distributed stochastic gradient tracking method in the non-convex setting with a one-point gradient estimator (1P-DSGT-NC). In the GT technique, every agent $i$ must keep and tend to two variables, a local copy $x_i\in\mathbb{R}^d$ of the decision variable and another auxiliary variable $y_i\in\mathbb{R}^d$. GT entails one ‘gradient descent’ step and one auxiliary variable update step. The reason for the technique’s name is that the auxiliary variable update usually allows it to track the gradient of the global objective function in the network. The gradient descent step updates the local optimization variables using the auxiliary variable, which itself tracks the global gradient. As shown in Algorithm \ref{alg:example}, $y_i$ is the sum of its previous iteration and the difference between the new and the old gradient estimates so that the update may include only the new information brought by the new estimate. $w_{ij}$ is the weight of the link between agents $i$ and $j$, so finally, the algorithm evolves by considering a weighted sum of neighboring updates and never sharing the local objective function nor its gradient (estimate).
Moreover, we assume that every local function is influenced by a stochastic variable $\xi_i\in\mathbb{R}^m$, and its query is subject to an additive scalar noise $\zeta_i$. At iteration $k$, the variables are denoted as $x_{i,k}$, $y_{i,k}$, $\xi_{i,k}$, and $\zeta_{i,k}$. 	

By querying the function once per algorithm iteration $k\in\mathbb{N}$, agent $i$ assembles its one-point gradient estimator $g_{i,k}$:

\begin{equation}\label{grdt_estimate}
	\begin{split}
		g_{i,k}&=z_{i,k}\tilde{f}_i(x_{i,k}+\gamma_{k}z_{i,k}, \xi_{i,k})\\
		&=z_{i,k}(f_i(x_{i,k}+\gamma_{k}z_{i,k}, \xi_{i,k})+\zeta_{i,k}),
	\end{split}
\end{equation}
where $z_{i,k} \in \mathbb{R}^d$ is a random perturbation vector generated by agent $i$, and $\gamma_{k}>0$ is a vanishing step size. We let $\eta_k>0$ be another vanishing step size, and we summarize the algorithm updates in Algorithm \ref{alg:example}. We note that at every iteration, every agent $i\in\mathcal{N}$ updates its variables independently and in parallel with all other agents in the network. 

\begin{algorithm}[tb]
	\caption{The 1P-DSGT-NC Algorithm}
	\label{alg:example}
	{\bfseries Input:} Initial arbitrary value $x_{i,0}\in\mathbb{R}^d$, the agents' coupling matrix $W$,  the initial step-sizes $\eta_0$ and $\gamma_{0}$, and setting $y_{i,0} = g_{i,0}$ 
	\begin{algorithmic}[1]
		\FOR{$k=0$ {\bfseries to} $K-1$, in parallel on all agents $i$,}
		\STATE Send $x_{i,k}$ and $y_{i,k}$ to neighbors
		\STATE Update the decision variable: \\
		$x_{i,k+1}=\sum_{j=1}^{n}w_{ij}(x_{j,k}-\eta_k y_{j,k})$
		\STATE Sample $z_{i,k+1}$ and update $\eta_{k+1}$ and $\gamma_{k+1}$
		\STATE Query the local function $f_i$ at the point: \\ 
		$x_{i,k+1}+\gamma_{k+1}z_{i,k+1}$
		\STATE Assemble $g_{i,k+1}$ according to (\ref{grdt_estimate})
		\STATE Update the auxiliary variable: \\ $y_{i,k+1}=\sum_{j=1}^{n}w_{ij}y_{j,k}+g_{i,k+1}-g_{i,k}$
		\ENDFOR 
	\end{algorithmic}
\end{algorithm}

We next let $\mathbf{x}_k:=[x_{1,k}, x_{2,k}, \ldots , x_{n,k}]^T$ and $\mathbf{y}_k:=[y_{1,k}, y_{2,k}, \ldots, y_{n,k}]^T$ $\in\mathbb{R}^{n\times d}$, and their means $\bar{x}_k:=\frac{1}{n}\mathbf{1}^T \mathbf{x}_k$ and $\bar{y}_k:=\frac{1}{n}\mathbf{1}^T \mathbf{y}_k$ $\in\mathbb{R}^{1\times d}$, respectively. The perturbation matrix is defined as $\mathbf{z}_{k}:=[z_{1,k},z_{2,k},\ldots,z_{n,k}]^T$, all $\in\mathbb{R}^{n\times d}$. Denoting the concatenated stochastic variables as $\mathbf{\xi}_k:= [\xi_{1,k},\xi_{2,k},\ldots,\xi_{n,k}]^T \in\mathbb{R}^{n\times m}$, then $\mathbf{g}_k:=g(\mathbf{x}_k,\mathbf{\xi}_k):=[g_1(x_{1,k},\xi_{1,k}),g_2(x_{2,k},\xi_{2,k}),\ldots,g_n(x_{n,k},\xi_{n,k})]^T\in\mathbb{R}^{n\times d}$ denotes the concatenated gradient estimate of all agents and its mean is $\bar{g} := \frac{1}{n}\mathbf{1}^T \mathbf{g}\in\mathbb{R}^{1\times d}$. 

The algorithm's updates can thus be written in matrix notation:
\begin{equation}\label{compact}
	\begin{split}
		&\mathbf{x}_{k+1}=W(\mathbf{x}_k-\eta_k \mathbf{y}_k)\\
		&\text{Query at: } \mathbf{x}_{k+1} + \gamma_{k+1}\mathbf{z}_{k+1}\\
		&\mathbf{y}_{k+1}=W\mathbf{y}_k+\mathbf{g}_{k+1}-\mathbf{g}_k,
	\end{split}
\end{equation}
with
\begin{equation}\label{GT-property}
	\bar{y}_k = \bar{g}_k \;\;\text{and}\;\; \bar{x}_{k+1} = \bar{x}_{k}-\eta_k \bar{g}_k,
\end{equation}
where the first equality in (\ref{GT-property}) is by $\mathbf{y}_{0}=\mathbf{g}_{0}$ and the recursion of the algorithm's updates, and the second equality is due to the doubly stochastic property of $W$.

\begin{remark}
	We only use matrix notation to simplify the analysis. In fact, all agents perform the update $x_{i,k+1}=\sum_{j=1}^{n}w_{ij}(x_{j,k}-\eta_k y_{j,k})$ in parallel, so the per-iteration cost is $O(nd)$ (as $w_{ij}$ is a scalar and both $x_{j,k}$ and $y_{j,k}$ are of dimension $d$). Same for step $7$ in Algorithm \ref{alg:example}. The sparsity of matrix $W$ helps the complexity by considering at most $m<n$ neighbors for any agent $i$, then we obtain $O(md)$.
\end{remark}

\section{Convergence Result} \label{cv-sec}	
In this section, we present the convergence results of Algorithm \ref{alg:example}. We begin by introducing necessary assumptions on the algorithm parameters.

\begin{assumption}\label{step_sizes}
	(the step-sizes) $\eta_k$ and $\gamma_k$ are two decreasing step sizes that converge to $0$ as $k\rightarrow\infty$ such that
	\begin{equation*}
		\sum_{k=1}^{\infty} \eta_k \gamma_k =\infty, \;\sum_{k=1}^{\infty} \eta_k\gamma_k^3<\infty, \;\text{and}\; \sum_{k=1}^{\infty} \eta_k^2<\infty.
	\end{equation*}
\end{assumption}
An example of the step sizes satisfying Assumption \ref{step_sizes} is the following form:
\begin{equation}\label{step-sizes-form}
	\eta_k = \eta_0(1+k)^{-\upsilon_1} \;\text{and}\; \gamma_k = \gamma_0 (1+k)^{-\upsilon_2}.
\end{equation}
Then, it's sufficient to find $\upsilon_1$ and $\upsilon_2$ such that $0<\upsilon_1+\upsilon_2\leq 1$, $\upsilon_1+3\upsilon_2>1$, and $\upsilon_1>0.5$.

\begin{assumption}\label{perturbation}
	(the random perturbation)
	The perturbation vector $z_{i,k} = (z_{i,k}^1, z_{i,k}^2, \ldots, z_{i,k}^d)^T$ has the same dimension as the local gradient. It is chosen independently by every agent $i\in\mathcal{N}$ from others and previous samples.
	Besides, it has i.i.d elements with 
	$\mathbb{E} (z_{i,k}^{d_j})^2 = \sigma_3>0$, $\forall {d_j}$, $\forall i$ and there exists a constant $\sigma_4 >0$ such that $\|z_{i,k}\|\leq \sigma_4$, $\forall i.$
\end{assumption}

To generate a perturbation vector satisfying Assumption \ref{perturbation}, we can choose every dimension of $z_{i,k}$ from the symmetric Bernoulli distribution on $\{-\frac{1}{\sqrt{d}},\frac{1}{\sqrt{d}}\}$. Then, we would have $\sigma_3=\frac{1}{d}$ and $\sigma_4=1$.

Let $\nabla F_i(x_i)\in\mathbb{R}^{1\times d}$ and  $\nabla^2 F_i(x_i)\in\mathbb{R}^{d\times d}$ denote the gradient of $F_i$ at the local variable and its Hessian matrix, respectively. Then, the gradient at $\mathbf{x}\in\mathbb{R}^{n\times d}$ is defined as
\begin{equation}
	\nabla F(\mathbf{x}):=[\nabla F_1(x_1),\nabla F_2(x_2),\ldots,\nabla F_n(x_n)]^T \in\mathbb{R}^{n\times d}.
\end{equation}
Let $h(\mathbf{x}):=\frac{1}{n}\mathbf{1}^T \nabla F(\mathbf{x}) \in\mathbb{R}^{1\times d}$ and denote by $\mathcal{H}_k=\{\mathbf{x}_0, \mathbf{y}_0, \mathbf{\xi}_0, \ldots, \mathbf{x}_{k-1}, \mathbf{y}_{k-1}, \mathbf{\xi}_{k-1}, \mathbf{x}_k\}$ the history sequence.
\begin{proposition}\label{prop} \cite{1, M-A}
	Let Assumptions \ref{noise} and \ref{perturbation} hold. Then, $g_{i,k}$ is a biased estimator of the agent's gradient $\nabla F_i(x_{i,k})$, $\forall i\in\mathcal{N}$, and 
	\begin{equation}
		\mathbb{E}_{z,\xi,\zeta} [g_{i,k}|\mathcal{H}_k] = \sigma_3 \gamma_{k}[\nabla F_i(x_{i,k})+ b_{i,k}],
	\end{equation}
	where $b_{i,k}$ denotes the bias with respect to the true gradient. Refer to Appendix \ref{gradient} for details.
\end{proposition}
\begin{lemma}\label{grdt_bnd}
	When Assumptions \ref{local_fcts}, \ref{noise}, and \ref{perturbation} are satisfied and	$\|\mathbf{x}_k\|<\infty$ almost surely, there exists a bounded constant $M > 0$, such that $E[\|\mathbf{g}_{k}\|^2|\mathcal{H}_k]< M$ almost surely.
	
	Proof: Refer to Appendix \ref{norm-sq}.
\end{lemma}

\begin{lemma}\label{lipschitz}
	By Assumption \ref{obj_fct}, we have
	\begin{equation}
		\|\nabla\mathcal{F}(\bar{x}_k)-h(\mathbf{x}_k)\|\leq \frac{L}{\sqrt{n}}\|\mathbf{x}_k-\mathbf{1}\bar{x}_k\|.
	\end{equation}
\end{lemma}
\begin{lemma}\label{rho_w} \cite{gr-tr1}
	Let Assumption \ref{network} hold, then 
	\begin{equation}
		\|W\omega-\mathbf{1}\bar{\omega}\| \leq\rho_w\|\omega-\mathbf{1}\bar{\omega}\|,\;\forall\omega \in \mathbb{R}^{n\times d},
	\end{equation}	
	where $\rho_w<1$ is the spectral norm of the matrix $W-\frac{1}{n}\mathbf{1}\mathbf{1}^T$ and $\bar{\omega} = \frac{1}{n}\mathbf{1}^T \omega$.
\end{lemma}

The main idea is to use the objective function's smoothness property in (\ref{smooth}). This property means that when the variable is varied, the gradient's variation does not shoot to infinity and stays confined within a scale of the variable's variation. We then study the variation employing the algorithm's properties in (\ref{GT-property}) and the stochastic error resulting from the gradient estimate vs its expectation. The variation becomes thus quantified by a step size multiplied by the average estimate. We use the facts that the expectation of estimate gives us biased access to the exact gradient (Proposition \ref{prop}), that both the norm squared of the estimate (Lemma \ref{grdt_bnd}) and the stochastic error are bounded. The smoothness inequality allows finding an upper bound on the exact gradient's norm where the key players are the step sizes: (updating $\gamma_k$) How far away from the variable do we need to go to query the function and find an estimate of the gradient at that variable, and (updating $\eta_k$) how far the algorithm should advance when updating a descent step? Answering these questions provides the right assumptions on the step sizes (Assumption \ref{step_sizes}). Finally, taking the telescoping sum of (\ref{smooth2}) provides the following technical result and justifies the summation assumptions on the step sizes.

\begin{theorem} \label{cv}
	Suppose there exists $\mathcal{F}^*:=\inf_{x\in\mathbb{R}^d}\mathcal{F}>-\infty$ and denote $\delta_k = \mathcal{F}(\bar{x}_{k})-\mathcal{F}^*$. Suppose also that $\|\mathbf{x}_k\|<\infty$ almost surely.
	Then, when Assumptions \ref{local_fcts}-\ref{network} and \ref{step_sizes}-\ref{perturbation} hold, the consensus error $\|\mathbf{x}_k-\mathbf{1}\bar{x}_k\|^2$ converges to $0$ as $k\rightarrow\infty$ and
	\begin{equation}
		\sum_{k=0}^{+\infty}\eta_k\gamma_k\|\nabla \mathcal{F}(\bar{x}_k)\|^2<\infty,
	\end{equation}
	which implies
	$\lim_{k\rightarrow\infty}\|\nabla \mathcal{F}(\bar{x}_k)\|=0$, almost surely.
	
	Proof: Refer to Appendix \ref{cv-proof}.
\end{theorem}
\section{Convergence Rate}	
We begin by defining the following constant terms:
\begin{equation}
	\begin{split}
		A_1 &= \frac{2}{\sigma_3\eta_0\gamma_0}\delta_0+\frac{4L^2}{n}\frac{1}{1-\rho_w^2}\|\mathbf{x}_0 - \mathbf{1}\bar{x}_0\|^2\\
		A_2 &= \frac{\sigma_4^6 \sigma_1^2}{2\sigma_3^2}\gamma_0^2(\upsilon_1+3\upsilon_2) \\
		A_3 &= 2\frac{L\bar{M}\eta_0}{\sigma_3\gamma_0}\upsilon_1\\
		A_4 &=\frac{12 L^2 G^2\eta_0^2}{n}\frac{\rho_w^2(1+\rho_w^2)}{(1-\rho_w^2)^2}\upsilon_1
	\end{split}
\end{equation}
where $\bar{M},G>0$ are defined such that the norm squared of the mean estimate $\|\bar{g}_k\|^2 \leq \bar{M}$ and the gradient tracking error $\|\mathbf{y}_{k}-\mathbf{1}\bar{y}_{k}\|^2 \leq G^2$ (refer to (\ref{g_bar}) in Appendix \ref{th-proof} and to (\ref{y_limit}) in Appendix \ref{xx_bar_sum_proof}, respectively).

As the convergence relies on the step sizes, we provide a form to the step sizes to locate an upper bound on the convergence rate. We present the resulting rate in the following theorem.
\begin{theorem}
	Let the step sizes satisfy (\ref{step-sizes-form}) with the further assumption $0<\upsilon_1+\upsilon_2<1$ and suppose $\|\mathbf{x}_k\|<\infty$ almost surely. When Assumptions \ref{local_fcts}-\ref{network} and \ref{step_sizes}-\ref{perturbation} are fulfilled,
	\begin{equation}
		\begin{split}
			&\frac{\sum_{k=0}^{K}\eta_k\gamma_k\mathbb{E}\big[\|\nabla \mathcal{F}(\bar{x}_k)\|^2\big]}{\sum_{k=0}^{K}\eta_k\gamma_k}\leq \frac{(1-\upsilon_1-\upsilon_2)}{ (K+2)^{1-\upsilon_1-\upsilon_2}-1}\\
			&\times\Bigg(A_1+\frac{A_2}{\upsilon_1+3\upsilon_2-1}+\frac{A_3}{2\upsilon_1-1}+\frac{A_4}{3\upsilon_1-1} \Bigg).
		\end{split}
	\end{equation}
	Proof: Refer to Appendix \ref{cv-rate}.
\end{theorem}
Optimizing for the time-varying part $O(\frac{1}{K^{1-\upsilon_1-\upsilon_2}})$, we find that the best choice for the exponents is $\upsilon_1=0.5$ and $\upsilon_2=\frac16$ for a rate of $O(\frac{1}{\sqrt[3]{K}})$. To avoid the constant term growing too large, one can take $\upsilon_1=\frac12+\frac{\epsilon}{2}$ and $\upsilon_2=\frac16+\frac{\epsilon}{2}$, for an arbitrarily small $\epsilon>0$, which achieve a rate $O\big(\frac{1}{K^{\frac13-\epsilon}}\big)$. We remark that when we use the perturbation vector example in Section \ref{cv-sec}, our bound scales as $O(d^2)$ in terms of the problem's dimension.

Another interesting aspect of our bound is that the employment of a biased one-point gradient estimate did not affect the dependence on the network size $n$ as compared with the first order-based gradient tracking method, for example, in \cite{nedic}. The only dependence on $n$ remains through the spectral norm $\rho_w$, as $G=O(\sqrt{n})$ and $\bar{M}=O(1)$ and the quantity $\frac{1}{n}\|\mathbf{x}_0 - \mathbf{1}\bar{x}_0\|^2 = \frac{1}{n}\sum_{i=1}^{n}\|x_{i,0} - x_{i,0}\|^2$ generally does not depend on $n$ as it's the average of the initial agents' decision variables. Thus, the graph topology determines the bounds' dependence on $n$.

\section{Numerical Example}\label{simulation}
This section presents a numerical example of our developed algorithm 1P-DSGT-NC to verify its efficacy.
We aim to classify $m$ images of two digits taken from the MNIST data set \cite{mnist} using logistic regression. These images labeled with $y=\{-1,+1\}$ are represented by $784$-dimensional vectors $X$, which are then compressed to dimension $d=10$ using a lossy autoencoder. They are split equally between $n$ agents, where every agent $i$ can only access their $m_i=\frac{m}{n}$ images. The collective objective is to locate $\theta\in\mathbb{R}^d$ that minimizes the following binary non-convex loss function: 
\begin{equation}
	\min_{\theta\in\mathbb{R}^d}\frac{1}{n}\sum_{i=1}^{n}\frac{1}{m}\sum_{j=1}^{m_i}\mathbb{E}_{u}\frac{1}{1+\exp(-u_{ij} y_{ij}.X_{ij}^T\theta)}+c\|\theta\|^2,
\end{equation}
where $u\sim\mathcal{N}(1,\sigma_u)$ is a random variable modeling stochastic variations on the querying of local functions. For example, unstable communications while fetching the data points from a local server. $c$ is the regularization constant.

\subsection{Simulation Setup}
A connected Erd\H{o}s-Rényi random graph with probability $0.3$ is used to model our agent network. We assemble the weight matrix $W$ using the Laplacian method \cite{gr-tr1}. Our algorithm is compared with a distributed stochastic gradient tracking method (DSGT-NC) based on an unbiased estimator with a bounded variance where the objective function is deterministic. This estimator is formed by calculating the exact gradient and adding noise. The final results are simulations averaged over $30$ instances.

We first classify the images of the digits $6$ and $7$. They are $m=12183$ images in total and divided equally over $n=31$ agents. Our parameters are generated as follows. The querying noise is $\zeta_{i,k}\sim\mathcal{N}(0,1)$, $\forall i\in\mathcal{N}$, the stochastic variable’s standard deviation is $\sigma_u = 0.01$, the regularization constant is $c=0.1$, the step sizes are $\eta_k = 1.5(k+1)^{-0.51}$ and $\gamma_{k} = 3.5(k+1)^{-0.17}$, and every dimension of the perturbation vector $z_k$ is chosen from $\{-\frac{1}{\sqrt{d}},\frac{1}{\sqrt{d}}\}$ with equal probability. For the DSGT-NC algorithm, the step size is $\eta_k = 2.5(k+1)^{-0.51}$, and no other noise than that on the exact gradient is considered. Both algorithms are initialized with the same random weights vectors $\theta_{i,0}\sim \mathcal{U}([-1,1]^d)$, $\forall i\in\mathcal{N}$, per simulation instance. 

We then classify images of digits $1$ and $2$, splitting the total $m=12700$ images over $n=50$ agents first and $n=100$ agents second and varying $\rho_w$ for $n=100$ agents to see its effect. When we kept the same model of Erd\H{o}s-Rényi, we obtained a smaller $\rho_w$ for a bigger $n$, i.e for $100$ agents, which makes sense for the same probability of edges. To obtain the same $\rho_w$ as for $50$ agents, we decreased the probability to $0.2$. All other parameters are left the same.

\subsection{Simulation Results}
\begin{figure}[ht]
	\vskip 0.2in
	\begin{center}
		\centerline{\includegraphics[width=\columnwidth]{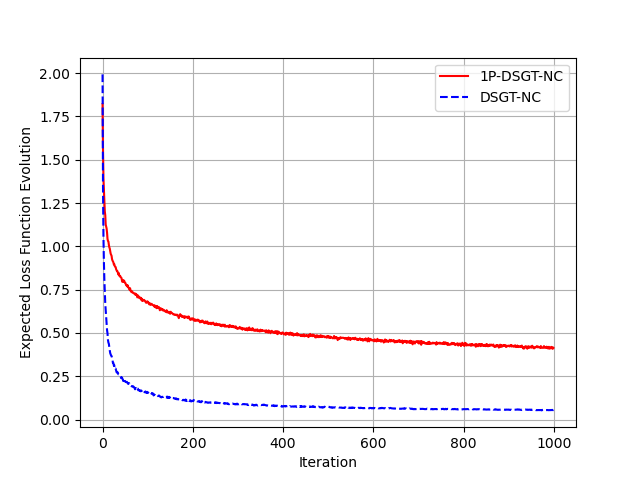}}
		\caption{The evolution of the expected loss function for the digits $6$ and $7$.}
		\label{loss-6-7}
	\end{center}
	\vskip -0.2in
\end{figure}
\begin{figure}[h!]
	\vskip 0.2in
	\begin{center}
		\centerline{\includegraphics[width=\columnwidth]{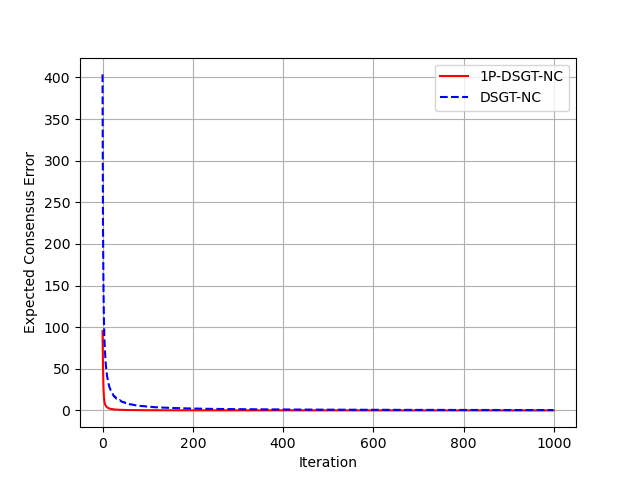}}
		\caption{The evolution of the consensus error for the digits $6$ and $7$.}
		\label{consensus-6-7}
	\end{center}
	\vskip -0.2in
\end{figure}

\begin{figure}[h!]
	\vskip 0.2in
	\begin{center}
		\centerline{\includegraphics[width=\columnwidth]{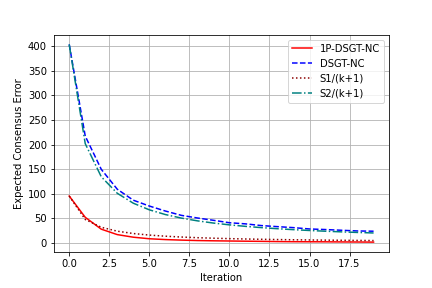}}
		\caption{The evolution of the consensus error for the digits $6$ and $7$ as compared with the rate $O(\frac{1}{K+1})$.}
		\label{zoom-consensus-6-7}
	\end{center}
	\vskip -0.2in
\end{figure}

\begin{figure}[h!]
	\vskip 0.2in
	\begin{center}
		\centerline{\includegraphics[width=\columnwidth]{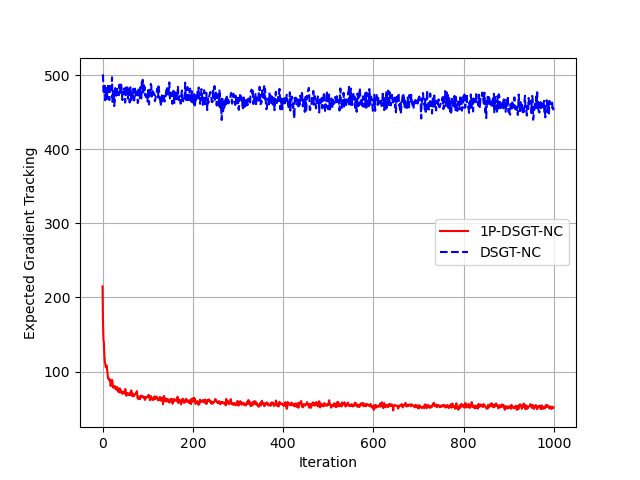}}
		\caption{The evolution of the gradient tracking error for the digits $6$ and $7$.}
		\label{tracking-6-7}
	\end{center}
	\vskip -0.2in
\end{figure}
\begin{figure}[h!]
	\vskip 0.2in
	\begin{center}
		\centerline{\includegraphics[width=\columnwidth]{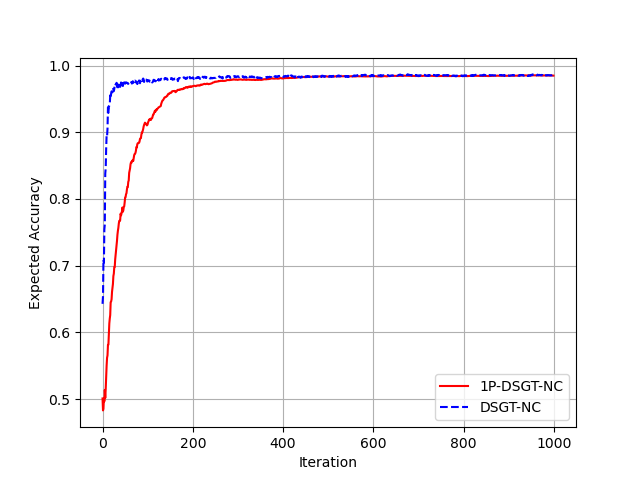}}
		\caption{The evolution of the accuracy for the digits $6$ and $7$.}
		\label{accuracy-6-7}
	\end{center}
	\vskip -0.2in
\end{figure}

\begin{figure}[h!]
	\vskip 0.2in
	\begin{center}
		\centerline{\includegraphics[width=\columnwidth]{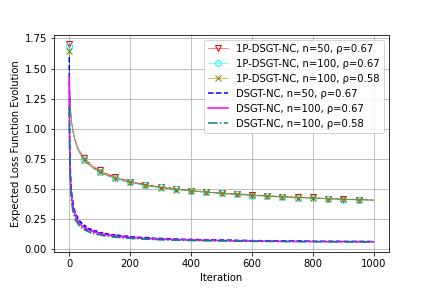}}
		\caption{The evolution of the expected loss function for the digits $1$ and $2$.}
		\label{loss-1-2}
	\end{center}
	\vskip -0.2in
\end{figure}

\begin{figure}[h!]
	\vskip 0.2in
	\begin{center}
		\centerline{\includegraphics[width=\columnwidth]{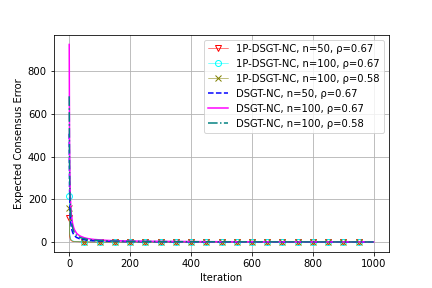}}
		\caption{The evolution of the consensus error for the digits $1$ and $2$.}
		\label{consensus-1-2}
	\end{center}
	\vskip -0.2in
\end{figure}

\begin{figure}[h!]
	\vskip 0.2in
	\begin{center}
		\centerline{\includegraphics[width=\columnwidth]{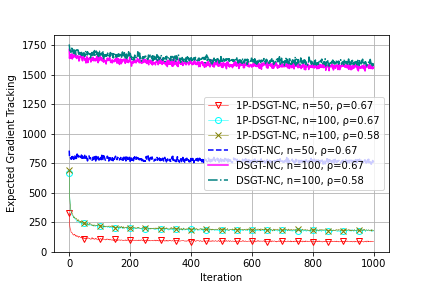}}
		\caption{The evolution of the gradient tracking error for the digits $1$ and $2$.}
		\label{tracking-1-2}
	\end{center}
	\vskip -0.2in
\end{figure}

\begin{figure}[h!]
	\vskip 0.2in
	\begin{center}
		\centerline{\includegraphics[width=\columnwidth]{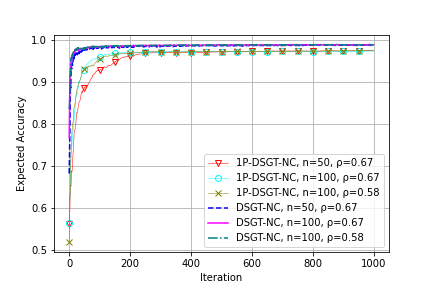}}
		\caption{The evolution of the accuracy for the digits $1$ and $2$.}
		\label{accuracy-1-2}
	\end{center}
	\vskip -0.2in
\end{figure}

\begin{figure}[h!]
	\vskip 0.2in
	\begin{center}
		\centerline{\includegraphics[width=\columnwidth]{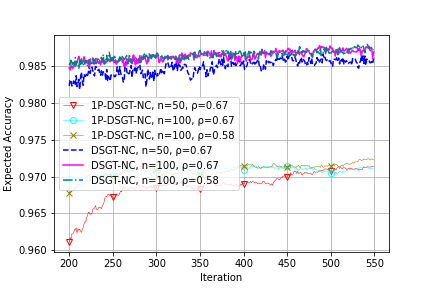}}
		\caption{Zooming on the evolution of the accuracy for the digits $1$ and $2$.}
		\label{zoom-accuracy-1-2}
	\end{center}
	\vskip -0.2in
\end{figure}

Figure \ref{loss-6-7} describes the development of the expected loss function for the digits $6$ and $7$ as the iterations advance. As FO methods naturally have better rates, it makes sense that DSGT-NC converges faster than our algorithm. However, the difference is still very small. On the other hand, the much faster convergence of the expected consensus error attained by our algorithm in Figure \ref{consensus-6-7} is surprising. To get a sense of the rate of convergence of this error, we add to the plot in Figure \ref{zoom-consensus-6-7} the two graphs $\frac{S_1}{k+1}$ and $\frac{S_2}{k+1}$, where $S_1=\mathbb{E}[\|\mathbf{x}_1-\mathbf{1}\bar{x}_1\|^2]$ the starting expected consensus error resulting from 1P-DSGT-NC and $S_2=\mathbb{E}[\|\mathbf{x}_1-\mathbf{1}\bar{x}_1\|^2]$ that resulting from DSGT-NC. We notice that 1P-DSGT-NC has a slightly better rate than $O(\frac{1}{K})$ and DSGT-NC a slightly worse rate.

Figure \ref{tracking-6-7} illustrates the expected gradient tracking error. While both this paper and that of \cite{nedic} prove that this error is bounded by constant and is linear in $n$, it could be that their constant is affected by other factors, and thus it's much greater.

In Figure \ref{accuracy-6-7}, we test the accuracy of both algorithms against an independent test set of size $1986$ images using the mean decision variable $\bar{\theta}_k =\frac{1}{n}\sum_{i=1}^{n}\theta_{i,k}$ as weight. The final accuracy achieved by our algorithm is $ 98.494461\%$ against $98.539778\%$ of the baseline. We conclude that, finally, the convergence rate did not impact the classification quality as both algorithms achieved excellent results.

Similarly, Figures \ref{loss-1-2}-\ref{zoom-accuracy-1-2} show the same results as previous ones. In Figure \ref{accuracy-1-2}, the test set is of size $2167$ images and the final accuracy with 1P-DSGT-NC is $97.303492\%$ for $(n,\rho_w)=(50,0.67)$, $97.285033\%$ for $(100, 0.67)$, and $97.340409\%$ for $(100,0.58)$. With DSGT-NC it is $98.681741\%$ for $(50, 0.67)$,  $98.777111\%$ for $(100, 0.67)$, and $98.780188\%$ for $(100, 0.58)$. We notice a slight improvement in the accuracy's convergence rate with the addition of agents for both algorithms in Figures \ref{accuracy-1-2} and \ref{zoom-accuracy-1-2}, despite keeping $\rho_w$ the same for one of the examples. The final performance generally stays the same for all cases, with the edge given to the smaller $\rho_w$. This is an interesting result as it proves that this algorithm is indeed scalable.

In Figure \ref{tracking-1-2}, the linear dependence on $n$ is clear for both algorithms, with a much better bound for 1P-DSGT-NC. 
\section{Conclusion}
In this paper, we have analyzed and demonstrated the convergence of the gradient tracking technique with non-convex objective functions that are subject to a stochastic process. Under the assumption that the gradient is not available and that only noisy single queries of the objective function are available at a time, we assembled a biased estimate of the gradient and proved a convergence rate of $O(\frac{1}{\sqrt[3]{K}})$ is possible. We then confirmed the efficacy of our algorithm with a binary logistic classification simulation model.

\bibliographystyle{apalike} 
\bibliography{1PDSGT}    

\newpage
\appendix
\onecolumn
\section{Estimated gradient}\label{gradient}
In this part, we elaborate on the properties of our gradient estimate. Denote by $\breve{g}_{i,k}=\mathbb{E}_{z,\xi,\zeta} [g_{i,k}|\mathcal{H}_k]$ the expected gradient estimate given the history sequence. Then,
\begin{equation}
	\begin{split}
		\breve{g}_{i,k}			&=\mathbb{E}_{z,\xi,\zeta} [z_{i,k}(f_i(x_{i,k}+\gamma_{k}z_{i,k}, \xi_{i,k})+\zeta_{i,k})|\mathcal{H}_k]\\
		&=\mathbb{E}_{z,\xi} [z_{i,k}f_i(x_{i,k}+\gamma_{k}z_{i,k}, \xi_{i,k})|\mathcal{H}_k]\\
		&=\mathbb{E}_{z} [z_{i,k}F_i(x_{i,k}+\gamma_{k}z_{i,k})|\mathcal{H}_k].
	\end{split}
\end{equation}
By Taylor’s theorem and the mean-valued theorem, there exists $\tilde{x}_{i,k} \in [x_{i,k}, x_{i,k}+\gamma_{k}z_{i,k}]$ such that
\begin{equation}
	F_i(x_{i,k}+\gamma_{k}z_{i,k})= F_i(x_{i,k})+\gamma_{k}\langle z_{i,k},\nabla F_i(x_{i,k})\rangle +\frac{\gamma_{k}^2}{2} \langle z_{i,k}, \nabla^2 F_i(\tilde{x}_{i,k})z_{i,k}\rangle. 
\end{equation}
Substituting in the expected gradient estimate,
\begin{equation}
	\begin{split}
		\breve{g}_{i,k} &= F_i(x_{i,k})\mathbb{E}_{z} [z_{i,k}]+\gamma_{k}\mathbb{E}_{z}[z_{i,k} z_{i,k}^T]\nabla F_i(x_{i,k})+ \frac{\gamma_{k}^2}{2} \mathbb{E}_{z}[z_{i,k} z_{i,k}^T \nabla^2  F_i(\tilde{x}_{i,k}) z_{i,k}|\mathcal{H}_k]\\
		&= \sigma_3 \gamma_{k}[\nabla F_i(x_{i,k})+ b_{i,k}],
	\end{split}
\end{equation}
where
\begin{equation}
	\begin{split}
		b_{i,k}&= \frac{\breve{g}_{i,k}}{\sigma_3 \gamma_{k}}-\nabla F_i(x_{i,k})\\
		&= \frac{\gamma_{k}}{2 \sigma_3 } \mathbb{E}_{z}[z_{i,k} z_{i,k}^T \nabla^2 F_i(\tilde{x}_{i,k}) z_{i,k}|\mathcal{H}_k].
	\end{split}
\end{equation}

By Assumptions \ref{obj_fct} and \ref{perturbation},
\begin{equation}\label{bias}
	\begin{split}
		\|b_{i,k}\| &\leq \frac{\gamma_{k}}{2 \sigma_3 } \mathbb{E}_{z}[ \| z_{i,k}\|_2 \|z_{i,k}^T\|_2 \|\nabla^2 F_i(\tilde{x}_{i,k})\|_2 \|z_{i,k}\|_2|\mathcal{H}_k] \\
		&\leq \gamma_{k} \frac{\sigma_4^3 \sigma_1}{2 \sigma_3}. 
	\end{split}
\end{equation}
Thus, the expected value of the mean gradient estimate is 
\begin{equation}\label{exp_g_bar}
	\begin{split}
		\tilde{g}_k &= \mathbb{E}[\bar{g}_k|\mathcal{H}_k]\\
		&= \frac{1}{n}\sum_{i=1}^{n} \mathbb{E}[g_{i,k}|\mathcal{H}_k]\\
		&= \frac{1}{n}\sum_{i=1}^{n}\sigma_3 \gamma_{k}[\nabla F_i(x_{i,k})+ b_{i,k}]\\
		&=\sigma_3 \gamma_{k}[h(\mathbf{x}_k)+ \bar{b}_{k}],
	\end{split}
\end{equation}
where the mean bias is
\begin{equation}\label{b_bar}
	\begin{split}
		\|\bar{b}_{k}\| &= \|\frac{1}{n}\sum_{i=1}^{n}b_{i,k}\|\\
		&\leq \frac{1}{n}\sum_{i=1}^{n}\|b_{i,k}\|\\
		&\leq \frac{1}{n}\sum_{i=1}^{n} \gamma_{k} \frac{\sigma_4^3 \sigma_1}{2 \sigma_3 }\\
		&=\gamma_{k} \frac{\sigma_4^3 \sigma_1}{2 \sigma_3 }.
	\end{split}	
\end{equation}

\subsection{Gradient Estimate Norm Squared Bound} \label{norm-sq}
We start by bounding the expected norm squared of the individual gradient estimates. $\forall i\in\mathcal{N}$, 
\begin{equation}\label{norm-sq-ineq}
	\begin{split}
		\mathbb{E}[\|g_{i,k}\|^2|\mathcal{H}_{k}]
		&=\mathbb{E}[\|z_{i,k}(f_i(x_{i,k}+\gamma_{k}z_{i,k}, \xi_{i,k})+\zeta_{i,k})\|^2|\mathcal{H}_{k}]\\
		&=\mathbb{E}[\|z_{i,k}\|^2\|f_i(x_{i,k}+\gamma_{k}z_{i,k}, \xi_{i,k})+\zeta_{i,k}\|^2|\mathcal{H}_{k}]\\
		&\overset{(a)}{\leq} \sigma_4^2 \mathbb{E}[(f_i(x_{i,k}+\gamma_{k}z_{i,k}, \xi_{i,k})+\zeta_{i,k})^2|\mathcal{H}_{k}]\\
		&\overset{(b)}{=}\sigma_4^2\mathbb{E}[f_i^2(x_{i,k}+\gamma_{k}z_{i,k}, \xi_{i,k})|\mathcal{H}_{k}]+\sigma_4^2\sigma_2\\
		&\overset{(c)}{\leq} \sigma_4^2\mathbb{E}[(\|f_i(0, \xi_{i,k})\|+L_{\xi_{i,k}}\|x_{i,k}+\gamma_{k}z_{i,k}\|)^2|\mathcal{H}_{k}]+\sigma_4^2\sigma_2\\
		&\overset{(d)}{\leq} 2\sigma_4^2\mathbb{E}[\mu_{\xi_{i,k}}^2+L_{\xi_{i,k}}^2(\|x_{i,k}\|+\gamma_{k}\sigma_4)^2|\mathcal{H}_{k}]+\sigma_4^2\sigma_2\\
		&\overset{(e)}{=} 2\sigma_4^2(\mu+L' (\|x_{i,k}\|+\gamma_{k}\sigma_4)^2)+\sigma_4^2\sigma_2\\
		&:= M_i \\
		&<\infty,
	\end{split}
\end{equation}
where $(a)$ is by Assumption \ref{perturbation}, $(b)$ Assumption \ref{noise}, $(c)$ Assumption \ref{local_fcts}. In $(d)$, we denote $\|f_i(0, \xi_{i,k})\|=\mu_{\xi_{i,k}}$ and the inequality is due to $\frac{x+y}{2}\leq \sqrt{\frac{x^2+y^2}{2}}$, $\forall x, y \in\mathbb{R}$. In $(e)$, $\mu=\mathbb{E}[\mu_{\xi_{i,k}}^2]$ and $L'=\mathbb{E}[L_{\xi_{i,k}}^2]$.

Then, we can write
\begin{equation}
	\mathbb{E}[\|\mathbf{g}_k\|^2|\mathcal{H}_{k}] =\sum_{i=1}^{n}\mathbb{E}[\|g_{i,k}\|^2|\mathcal{H}_{k}]< n \times \sup_{i\in\mathcal{N}} M_i := M<\infty.
\end{equation}

\section{Proof of Convergence}\label{cv-proof}
\begin{lemma}\label{xx_bar_sum}
	If all Assumptions \ref{local_fcts}, \ref{noise}, \ref{network}, \ref{step_sizes}, and \ref{perturbation} are fulfilled and the inequality $\|\mathbf{x}_k\|<\infty$ holds almost surely, then $\lim_{k\rightarrow\infty}\|\mathbf{x}_k-\mathbf{1}\bar{x}_k\|^2 =0 $.
	In addition, both sums
	\begin{equation}
		\sum_{k=0}^{\infty}\| \mathbf{x}_k-\mathbf{1}\bar{x}_k\|^2<\infty \;\text{and}\;\sum_{k=0}^{\infty}\gamma_{k}\eta_k\| \mathbf{x}_k-\mathbf{1}\bar{x}_k\|^2<\infty
	\end{equation}
	converge almost surely.
	
	Proof: See Appendix \ref{xx_bar_sum_proof}.
\end{lemma}
\subsection{Proof of Theorem \ref{cv}} \label{th-proof}
As a reminder of (\ref{GT-property}), $\bar{x}_{k+1} = \bar{x}_{k}-\eta_k \bar{g}_k$. By the $L$-smoothness of the objective function in Assumption (\ref{obj_fct}) and defining $e_k=\bar{g}_k-\mathbb{E}[\bar{g}_k|\mathcal{H}_k]$ as the stochastic error, we have
\begin{equation*}
	\begin{split}
		\mathcal{F}(\bar{x}_{k+1})
		&\leq \mathcal{F}(\bar{x}_k)-\eta_k\langle\nabla \mathcal{F}(\bar{x}_k), \bar{g}_k\rangle +\frac{\eta_k^2 L}{2}\|\bar{g}_k\|^2\\
		&= \mathcal{F}(\bar{x}_k)-\eta_k\langle\nabla \mathcal{F}(\bar{x}_k), e_k+\mathbb{E}[\bar{g}_k|\mathcal{H}_k]\rangle +\frac{\eta_k^2 L}{2}\|\bar{g}_k\|^2\\
		&\overset{(a)}{\leq} \mathcal{F}(\bar{x}_k)-\eta_k\langle\nabla \mathcal{F}(\bar{x}_k), e_k\rangle-\sigma_3\eta_k\gamma_k\langle\nabla \mathcal{F}(\bar{x}_k), h(\mathbf{x}_k)+\bar{b}_k\rangle +\frac{\eta_k^2 L}{2}\bar{M}\\
		&= \mathcal{F}(\bar{x}_k)-\eta_k\langle\nabla \mathcal{F}(\bar{x}_k), e_k\rangle-\sigma_3\eta_k\gamma_k\langle\nabla \mathcal{F}(\bar{x}_k), h(\mathbf{x}_k)+\bar{b}_k+\nabla\mathcal{F}(\bar{x}_k)-\nabla\mathcal{F}(\bar{x}_k)\rangle +\frac{L\bar{M}}{2}\eta_k^2 \\
		&= \mathcal{F}(\bar{x}_k)-\eta_k\langle\nabla \mathcal{F}(\bar{x}_k), e_k\rangle-\sigma_3\eta_k\gamma_k\|\nabla \mathcal{F}(\bar{x}_k)\|^2 -\sigma_3\eta_k\gamma_k\langle\nabla \mathcal{F}(\bar{x}_k),\bar{b}_k\rangle \\&+\sigma_3\eta_k\gamma_k\langle\nabla \mathcal{F}(\bar{x}_k), \nabla\mathcal{F}(\bar{x}_k)-h(\mathbf{x}_k)\rangle +\frac{L\bar{M}}{2}\eta_k^2 \\
	\end{split}
\end{equation*}
\begin{equation}\label{smo}
	\begin{split}
		&\overset{(b)}{\leq} \mathcal{F}(\bar{x}_k)-\eta_k\langle\nabla \mathcal{F}(\bar{x}_k), e_k\rangle-\sigma_3\eta_k\gamma_k\|\nabla \mathcal{F}(\bar{x}_k)\|^2 +\sigma_3\eta_k\gamma_k\|\nabla \mathcal{F}(\bar{x}_k)\|\|\bar{b}_k\| \\
		&+\sigma_3\eta_k\gamma_k\|\nabla \mathcal{F}(\bar{x}_k)\|\|\nabla\mathcal{F}(\bar{x}_k)-h(\mathbf{x}_k)\| +\frac{L\bar{M}}{2}\eta_k^2 \\
		&\overset{(c)}{\leq} \mathcal{F}(\bar{x}_k)-\eta_k\langle\nabla \mathcal{F}(\bar{x}_k), e_k\rangle-\sigma_3\eta_k\gamma_k\|\nabla \mathcal{F}(\bar{x}_k)\|^2 +\frac{\sigma_3\eta_k\gamma_k}{4}\|\nabla \mathcal{F}(\bar{x}_k)\|^2\\
		&+\sigma_3\eta_k\gamma_k\|\bar{b}_k\|^2 +\frac{\sigma_3\eta_k\gamma_k}{4}\|\nabla \mathcal{F}(\bar{x}_k)\|^2+\sigma_3\eta_k\gamma_k\|\nabla\mathcal{F}(\bar{x}_k)-h(\mathbf{x}_k)\|^2 +\frac{L\bar{M}}{2}\eta_k^2 \\
		&= \mathcal{F}(\bar{x}_k)-\eta_k\langle\nabla \mathcal{F}(\bar{x}_k), e_k\rangle-\frac{\sigma_3\eta_k\gamma_k}{2}\|\nabla \mathcal{F}(\bar{x}_k)\|^2 +\sigma_3\eta_k\gamma_k\|\bar{b}_k\|^2 +\sigma_3\eta_k\gamma_k\|\nabla\mathcal{F}(\bar{x}_k)-h(\mathbf{x}_k)\|^2 +\frac{L\bar{M}}{2}\eta_k^2 \\
		&\overset{(d)}{\leq} \mathcal{F}(\bar{x}_k)-\eta_k\langle\nabla \mathcal{F}(\bar{x}_k), e_k\rangle-\frac{\sigma_3\eta_k\gamma_k}{2}\|\nabla \mathcal{F}(\bar{x}_k)\|^2 + \frac{\sigma_4^6 \sigma_1^2}{4 \sigma_3}\eta_k\gamma_k^3 +\frac{\sigma_3L^2}{n}\eta_k\gamma_k\|\mathbf{x}_k-\mathbf{1}\bar{x}_k\|^2 +\frac{L\bar{M}}{2}\eta_k^2. \\
	\end{split}
\end{equation}
where $(a)$ is by (\ref{exp_g_bar}) and (\ref{g_bar}), $(b)$ is by Cauchy-Schwarz inequality, $(c)$ is by $-2 \epsilon \times \frac{1}{\epsilon}\langle a,b\rangle = -2 \langle \epsilon a,\frac{1}{\epsilon} b\rangle\leq \epsilon^2\|a\|^2 +\frac{1}{\epsilon^2}\|b\|^2$, and $(d)$ is by (\ref{b_bar}) and Lemma \ref{lipschitz}.

By Lemma \ref{grdt_bnd},
\begin{equation}\label{g_bar}
	\|\bar{g}_k\|^2  =\|\frac{1}{n}\sum_{i=1}^{n}g_{i,k}\|^2\leq\frac{n}{n^2}\sum_{i=1}^{n}\|g_{i,k}\|^2 = \frac{1}{n}\sum_{i=1}^{n}\|g_{i,k}\|^2\leq\bar{M}<\infty, \;\text{almost surely.}
\end{equation}

By taking the telescoping sum of (\ref{smo}),
\begin{equation}
	\begin{split}
		\mathcal{F}(\bar{x}_{K+1})
		&\leq \mathcal{F}(\bar{x}_0)-\sum_{k=0}^{K}\eta_k\langle\nabla \mathcal{F}(\bar{x}_k), e_k\rangle-\frac{\sigma_3}{2}\sum_{k=0}^{K}\eta_k\gamma_k\|\nabla \mathcal{F}(\bar{x}_k)\|^2 + \frac{\sigma_4^6 \sigma_1^2}{4 \sigma_3}\sum_{k=0}^{K}\eta_k\gamma_k^3 +\frac{\sigma_3L^2}{n}\sum_{k=0}^{K}\eta_k\gamma_k\|\mathbf{x}_k-\mathbf{1}\bar{x}_k\|^2\\ &+\frac{L\bar{M}}{2}\sum_{k=0}^{K}\eta_k^2. \\
	\end{split}
\end{equation}
We know that $\delta_k = \mathcal{F}(\bar{x}_{k})-\mathcal{F}^*>0$ by definition. Then, 
\begin{equation}
	\begin{split}
		0 \leq\delta_{K+1}
		\leq &\delta_0-\sum_{k=0}^{K}\eta_k\langle\nabla \mathcal{F}(\bar{x}_k), e_k\rangle-\frac{\sigma_3}{2}\sum_{k=0}^{K}\eta_k\gamma_k\|\nabla \mathcal{F}(\bar{x}_k)\|^2 +\frac{\sigma_4^6 \sigma_1^2}{4 \sigma_3}\sum_{k=0}^{K}\eta_k\gamma_k^3 +\frac{\sigma_3L^2}{n}\sum_{k=0}^{K}\eta_k\gamma_k\|\mathbf{x}_k-\mathbf{1}\bar{x}_k\|^2 \\&+\frac{L\bar{M}}{2}\sum_{k=0}^{K}\eta_k^2.\\
	\end{split}
\end{equation}
Finally,
\begin{equation}\label{grdt_ineq}
	\sum_{k=0}^{K}\eta_k\gamma_k\|\nabla \mathcal{F}(\bar{x}_k)\|^2 \leq\frac{2}{\sigma_3}\delta_0-\frac{2}{\sigma_3}\sum_{k=0}^{K}\eta_k\langle\nabla \mathcal{F}(\bar{x}_k), e_k\rangle+ \frac{\sigma_4^6 \sigma_1^2}{2 \sigma_3^2}\sum_{k=0}^{K}\eta_k\gamma_k^3 +\frac{2L^2}{n}\sum_{k=0}^{K}\eta_k\gamma_k\|\mathbf{x}_k-\mathbf{1}\bar{x}_k\|^2 +\frac{L\bar{M}}{\sigma_3}\sum_{k=0}^{K}\eta_k^2.
\end{equation}
By Assumption \ref{obj_fct}, $\|\nabla \mathcal{F}(\bar{x}_k)\|$ is bounded for any $\bar{x}_k\in\mathbb{R}^d$. As one can easily see that the sequence $\{\sum_{k=K}^{K'} \eta_k e_k\}_{K'\geq K}$ is a martingale, we then use Doob's martingale inequality \cite{doob} to prove that the norm of this sequence is bounded for any $K'\geq K$, which results in

\begin{equation}
	\lim_{K\rightarrow\infty}\|\sum_{k=0}^{K}\eta_k\langle\nabla \mathcal{F}(\bar{x}_k),e_k\rangle\|<\infty.
\end{equation} 
And since as $K\rightarrow\infty$, by Assumption \ref{step_sizes}, $\sum_k\eta_k\gamma_k^3$ and $\sum_k\eta_k^2$ converge, and by Lemma \ref{xx_bar_sum} $\sum_k\eta_k\gamma_k\|\mathbf{x}_k-\mathbf{1}\bar{x}_k\|^2$ converges, then $\sum_k\eta_k\gamma_k\|\nabla \mathcal{F}(\bar{x}_k)\|^2$ must converge as well.

As $\sum_k\eta_k\gamma_k$ diverges, this implies that $\lim_{k\rightarrow\infty}\inf \|\nabla\mathcal{F}(\bar{x}_k)\|=0$. 

Next, let $\epsilon>0$ be an arbitrary value, and consider the event
\begin{equation*}
	A_{\epsilon}:=\biggl\{\lim_{k\rightarrow\infty}\sup \|\nabla\mathcal{F}(\bar{x}_k)\|\geq \epsilon\biggr\}.
\end{equation*}
When event $A_{\epsilon}$ is true, an arbitrary subsequence $\big(\|\nabla\mathcal{F}(\bar{x}_{k_l})\|\big)_{l\in\mathbb{N}}$ of $\|\nabla\mathcal{F}(\bar{x}_k)\|$ can always be found, such that $\|\nabla\mathcal{F}(\bar{x}_{k_l})\|\geq \epsilon - \nu$, $\forall l$, for $\epsilon - \nu>0$ and $\nu>0$.

Then,
\begin{equation}
	\begin{split}
		\|\nabla\mathcal{F}(\bar{x}_{k_l+1})\|&\geq \|\nabla\mathcal{F}(\bar{x}_{k_l})\|-\|\nabla\mathcal{F}(\bar{x}_{k_l+1})-\nabla\mathcal{F}(\bar{x}_{k_l})\| \\
		&\geq \epsilon - \nu-L\|\bar{x}_{k_l+1}-\bar{x}_{k_l}\|\\
		&= \epsilon - \nu-L\eta_{k_l}\|\bar{g}_{k_l}\|\\
		&\geq \epsilon - \nu-L\bar{M}\eta_{k_l},
	\end{split}
\end{equation}
Since $k_l\rightarrow\infty$ as $l\rightarrow\infty$, we can find a subsequence of $(k_{l_p})_{p\in\mathbb{N}}$ such that $k_{l_{p+1}}-k_{l_p}>1$. Thus,
\begin{equation}
	\begin{split}
		\sum_{k=0}^{\infty} \eta_{k+1}\gamma_{k+1}\|\nabla \mathcal{F}(\bar{x}_{k+1})\|^2 &\geq (\epsilon - \nu)^2\sum_{k=0}^{\infty}\eta_{k+1}\gamma_{k+1}-2(\epsilon - \nu)L\bar{M}\sum_{k=0}^{\infty}\eta_{k+1}\gamma_{k+1}\eta_k+L^2\bar{M}^2\sum_{k=0}^{\infty}\eta_{k+1}\gamma_{k+1}\eta_{k}^2 \\
		&\geq (\epsilon - \nu)^2\sum_{k=0}^{\infty}\eta_{k+1}\gamma_{k+1}-2(\epsilon - \nu)L\bar{M}\sum_{k=0}^{\infty}\eta_{k}^2\gamma_{k}+L^2\bar{M}^2\sum_{k=0}^{\infty}\eta_{k+1}\gamma_{k+1}\eta_{k}^2 \\\\
		&=+\infty,
	\end{split}
\end{equation}
as the first series diverges and the second and the third converge by Assumption \ref{step_sizes}.
This implies that on $A_{\epsilon}$ the series $\sum_{k} \eta_{k}\gamma_{k}\|\nabla \mathcal{F}(\bar{x}_{k})\|^2$ diverges. This is a contradiction as this series converges almost surely by (\ref{grdt_ineq}). Therefore, $\mathbb{P}(A_{\epsilon})=0$ and as a result,
\begin{equation}
	\mathbb{P}\bigg(\lim_{k\rightarrow\infty}\|\nabla\mathcal{F}(\bar{x}_{k})\|>0\bigg) = \mathbb{P}\bigg(\bigcup_{\epsilon>0}A_{\epsilon}\bigg)=0.
\end{equation}
We conclude that $\nabla\mathcal{F}(\bar{x}_{k})$ converges almost surely.

\section{Rate of Convergence} \label{cv-rate}
By the $L$-smoothness, 
\begin{equation}
	\mathcal{F}(\bar{x}_{k+1})
	\leq \mathcal{F}(\bar{x}_k)-\eta_k\langle\nabla \mathcal{F}(\bar{x}_k), \bar{g}_k\rangle +\frac{\eta_k^2 L}{2}\|\bar{g}_k\|^2.
\end{equation}
Taking the conditional expectation given the history sequence $\mathcal{H}_k$ on both sides,
\begin{equation*}
	\begin{split}
		\mathbb{E}[\mathcal{F}(\bar{x}_{k+1})|\mathcal{H}_k]
		&\overset{(a)}{\leq} \mathcal{F}(\bar{x}_k)-\sigma_3\eta_k\gamma_k\langle\nabla \mathcal{F}(\bar{x}_k), h(\mathbf{x}_k)+\bar{b}_k\rangle +\frac{\eta_k^2 L}{2}\bar{M}\\
		&= \mathcal{F}(\bar{x}_k)-\sigma_3\eta_k\gamma_k\langle\nabla \mathcal{F}(\bar{x}_k), h(\mathbf{x}_k)+\bar{b}_k+\nabla\mathcal{F}(\bar{x}_k)-\nabla\mathcal{F}(\bar{x}_k)\rangle +\frac{L\bar{M}}{2}\eta_k^2 \\
		&= \mathcal{F}(\bar{x}_k)-\sigma_3\eta_k\gamma_k\|\nabla \mathcal{F}(\bar{x}_k)\|^2 -\sigma_3\eta_k\gamma_k\langle\nabla \mathcal{F}(\bar{x}_k),\bar{b}_k\rangle \\&+\sigma_3\eta_k\gamma_k\langle\nabla \mathcal{F}(\bar{x}_k), \nabla\mathcal{F}(\bar{x}_k)-h(\mathbf{x}_k)\rangle +\frac{L\bar{M}}{2}\eta_k^2 \\
		&\overset{(b)}{\leq} \mathcal{F}(\bar{x}_k)-\sigma_3\eta_k\gamma_k\|\nabla \mathcal{F}(\bar{x}_k)\|^2 +\sigma_3\eta_k\gamma_k\|\nabla \mathcal{F}(\bar{x}_k)\|\|\bar{b}_k\| \\
		&+\sigma_3\eta_k\gamma_k\|\nabla \mathcal{F}(\bar{x}_k)\|\|\nabla\mathcal{F}(\bar{x}_k)-h(\mathbf{x}_k)\| +\frac{L\bar{M}}{2}\eta_k^2 \\
		&\overset{(c)}{\leq} \mathcal{F}(\bar{x}_k)-\sigma_3\eta_k\gamma_k\|\nabla \mathcal{F}(\bar{x}_k)\|^2 +\frac{\sigma_3\eta_k\gamma_k}{4}\|\nabla \mathcal{F}(\bar{x}_k)\|^2\\
		&+\sigma_3\eta_k\gamma_k\|\bar{b}_k\|^2 +\frac{\sigma_3\eta_k\gamma_k}{4}\|\nabla \mathcal{F}(\bar{x}_k)\|^2+\sigma_3\eta_k\gamma_k\|\nabla\mathcal{F}(\bar{x}_k)-h(\mathbf{x}_k)\|^2 +\frac{L\bar{M}}{2}\eta_k^2 \\
		&= \mathcal{F}(\bar{x}_k)-\frac{\sigma_3\eta_k\gamma_k}{2}\|\nabla \mathcal{F}(\bar{x}_k)\|^2 +\sigma_3\eta_k\gamma_k\|\bar{b}_k\|^2 +\sigma_3\eta_k\gamma_k\|\nabla\mathcal{F}(\bar{x}_k)-h(\mathbf{x}_k)\|^2 +\frac{L\bar{M}}{2}\eta_k^2 \\
\end{split}
\end{equation*}
\begin{equation}\label{smo2}
	\begin{split}
		&\overset{(d)}{\leq} \mathcal{F}(\bar{x}_k)-\frac{\sigma_3\eta_k\gamma_k}{2}\|\nabla \mathcal{F}(\bar{x}_k)\|^2 + \frac{\sigma_4^6 \sigma_1^2}{4 \sigma_3}\eta_k\gamma_k^3 +\frac{\sigma_3L^2}{n}\eta_k\gamma_k\|\mathbf{x}_k-\mathbf{1}\bar{x}_k\|^2 +\frac{L\bar{M}}{2}\eta_k^2. \\
	\end{split}
\end{equation}
where $(a)$ is by (\ref{exp_g_bar}) and (\ref{g_bar}), $(b)$ is by Cauchy-Schwarz inequality, $(c)$ is by $-2 \epsilon \times \frac{1}{\epsilon}\langle a,b\rangle = -2 \langle \epsilon a,\frac{1}{\epsilon} b\rangle\leq \epsilon^2\|a\|^2 +\frac{1}{\epsilon^2}\|b\|^2$, and $(d)$ is by (\ref{b_bar}) and Lemma \ref{lipschitz}.
By taking the full expectation then the telescoping sum of (\ref{smo2})

\begin{equation}
	\begin{split}
		\mathbb{E}[\mathcal{F}(\bar{x}_{K+1})]
		\leq &\mathcal{F}(\bar{x}_0)-\frac{\sigma_3}{2}\sum_{k=0}^{K}\eta_k\gamma_k \mathbb{E}[\|\nabla \mathcal{F}(\bar{x}_k)\|^2]+ \frac{\sigma_4^6 \sigma_1^2}{4 \sigma_3}\sum_{k=0}^{K}\eta_k\gamma_k^3 +\frac{\sigma_3L^2}{n}\sum_{k=0}^{K}\eta_k\gamma_k\mathbb{E}[\|\mathbf{x}_k-\mathbf{1}\bar{x}_k\|^2]+\frac{L\bar{M}}{2}\sum_{k=0}^{K}\eta_k^2. \\
	\end{split}
\end{equation}
Then,
\begin{equation}
	\begin{split}
		0\leq\mathbb{E}[\delta_{K+1}]
		\leq &\delta_0-\frac{\sigma_3}{2}\sum_{k=0}^{K}\eta_k\gamma_k \mathbb{E}[\|\nabla \mathcal{F}(\bar{x}_k)\|^2]+ \frac{\sigma_4^6 \sigma_1^2}{4 \sigma_3}\sum_{k=0}^{K}\eta_k\gamma_k^3 +\frac{\sigma_3L^2}{n}\sum_{k=0}^{K}\eta_k\gamma_k\mathbb{E}[\|\mathbf{x}_k-\mathbf{1}\bar{x}_k\|^2]+\frac{L\bar{M}}{2}\sum_{k=0}^{K}\eta_k^2. \\
	\end{split}
\end{equation}
Finally, 
\begin{equation}\label{smo3}
	\begin{split}
		\sum_{k=0}^{K}\eta_k\gamma_k \mathbb{E}[\|\nabla \mathcal{F}(\bar{x}_k)\|^2]
		\leq \frac{2}{\sigma_3}\delta_0+ \frac{\sigma_4^6 \sigma_1^2}{2 \sigma_3^2}\sum_{k=0}^{K}\eta_k\gamma_k^3 +\frac{2L^2}{n}\sum_{k=0}^{K}\eta_k\gamma_k\mathbb{E}[\|\mathbf{x}_k-\mathbf{1}\bar{x}_k\|^2]+\frac{L\bar{M}}{\sigma_3}\sum_{k=0}^{K}\eta_k^2. \\
	\end{split}
\end{equation}

Let the step sizes satisfy (\ref{step-sizes-form}).	We know that, $\forall K>0$, 
\begin{equation}
	\begin{split}
		\sum_{k=0}^{K}\eta_k\gamma_k^3 = \eta_0 \gamma_0^3 +\sum_{k=1}^{K}\eta_k\gamma_k^3 &\leq \eta_0 \gamma_0^3\bigg(1 + \int_{0}^{K}(x+1)^{-\upsilon_1-3\upsilon_2}dx\bigg)\\
		&= \eta_0 \gamma_0^3\bigg(1 + \frac{1}{\upsilon_1+3\upsilon_2-1}-\frac{(K+1)^{-\upsilon_1-3\upsilon_2+1}}{\upsilon_1+3\upsilon_2-1}\bigg)\\
		&\leq \eta_0 \gamma_0^3\bigg(1 + \frac{1}{\upsilon_1+3\upsilon_2-1}\bigg)\\
		&= \eta_0 \gamma_0^3\bigg(\frac{\upsilon_1+3\upsilon_2}{\upsilon_1+3\upsilon_2-1}\bigg).\\\
	\end{split}
\end{equation}	
Similarly,
\begin{equation}
	\sum_{k=0}^{K}\eta_k^2 \leq \eta_0^2 \bigg(\frac{2\upsilon_1}{2\upsilon_1-1}\bigg) \;\;\text{and}\;\; 		\sum_{k=0}^{K}\eta_k^3 \leq \eta_0^3 \bigg(\frac{3\upsilon_1}{3\upsilon_1-1}\bigg).
\end{equation}
Next, when $0<\upsilon_1+\upsilon_2<1$,
\begin{equation}
	\begin{split}
		\sum_{k=0}^{K}\eta_k\gamma_k &\geq \eta_0\gamma_0\int_{0}^{K+1} (x+1)^{-\upsilon_1-\upsilon_2} dx\\
		&= \frac{\eta_0\gamma_0}{(1-\upsilon_1-\upsilon_2)} \bigg((K+2)^{1-\upsilon_1-\upsilon_2}-1\bigg).
	\end{split}
\end{equation}
Then, substituting back in (\ref{smo3}),
\begin{equation}
	\begin{split}
		\frac{\sum_{k=0}^{K}\eta_k\gamma_k\mathbb{E}[\|\nabla \mathcal{F}(\bar{x}_k)\|^2]}{\sum_{k=0}^{K}\eta_k\gamma_k} \leq\frac{(1-\upsilon_1-\upsilon_2)}{ (K+2)^{1-\upsilon_1-\upsilon_2}-1}\Bigg(&\frac{2}{\sigma_3\eta_0\gamma_0}\delta_0+\frac{\sigma_4^6 \sigma_1^2}{2 \sigma_3^2}\gamma_0^2\frac{\upsilon_1+3\upsilon_2}{\upsilon_1+3\upsilon_2-1}+\frac{L\bar{M}\eta_0}{\sigma_3\gamma_0}\frac{2\upsilon_1}{2\upsilon_1-1} \\
		&+\frac{4L^2}{n}\frac{1}{(1-\rho_w^2)}\|\mathbf{x}_0 - \mathbf{1}\bar{x}_0\|^2 +\frac{4L^2G^2\eta_0^2}{n }\frac{\rho_w^2(1+\rho_w^2)}{(1-\rho_w^2)^2}\frac{3\upsilon_1}{3\upsilon_1-1} \Bigg).
	\end{split}
\end{equation}
Refer to (\ref{bnd_sum_xx}) and (\ref{bnd_sum_xx2}) for the bounding of the term $\sum_k \gamma_k\eta_k\| \mathbf{x}_{k}-\mathbf{1}\bar{x}_{k}\|^2$.

\subsection{Proof of Lemma \ref{xx_bar_sum}}\label{xx_bar_sum_proof}
We substitute the variables by their updates in (\ref{compact}).
\begin{equation}\label{x_fct_previous}
	\begin{split}
		\| \mathbf{x}_{k+1}-\mathbf{1}\bar{x}_{k+1}\|^2 &= \| W \mathbf{x}_k -\eta_k W \mathbf{y}_k - \mathbf{1}\bar{x}_k +\eta_k \mathbf{1}\bar{y}_k\|^2 \\
		&= \| W \mathbf{x}_k - \mathbf{1}\bar{x}_k\|^2 -2\eta_k\langle W \mathbf{x}_k - \mathbf{1}\bar{x}_k, W \mathbf{y}_k-\mathbf{1}\bar{y}_k \rangle +\eta_k^2 \| W \mathbf{y}_k-\mathbf{1}\bar{y}_k \|^2 \\
		&\overset{(a)}{\leq} \| W \mathbf{x}_k - \mathbf{1}\bar{x}_k\|^2 + \eta_k[\frac{1-\rho_w^2}{2\rho_w^2\eta_k}\| W \mathbf{x}_k - \mathbf{1}\bar{x}_k\|^2+ \frac{2\rho_w^2\eta_k}{1-\rho_w^2}\| W \mathbf{y}_k-\mathbf{1}\bar{y}_k \|^2] \\
		&\hspace{0.5cm}+ \eta_k^2 \|W \mathbf{y}_k-\mathbf{1}\bar{y}_k \|^2\\
		&\overset{(b)}{\leq} \rho_w^2\| \mathbf{x}_k - \mathbf{1}\bar{x}_k\|^2 + \rho_w^2\eta_k[\frac{1-\rho_w^2}{2\rho_w^2\eta_k}\| \mathbf{x}_k - \mathbf{1}\bar{x}_k\|^2+ \frac{2\rho_w^2\eta_k}{1-\rho_w^2}\|\mathbf{y}_k-\mathbf{1}\bar{y}_k \|^2]  \\
		&\hspace{0.5cm}+\rho_w^2 \eta_k^2 \| \mathbf{y}_k-\mathbf{1}\bar{y}_k \|^2\\
		&= \frac{1+\rho_w^2}{2}\| \mathbf{x}_k - \mathbf{1}\bar{x}_k\|^2+\eta_k^2\frac{(1+\rho_w^2)\rho_w^2}{1-\rho_w^2}\| \mathbf{y}_k-\mathbf{1}\bar{y}_k \|^2,
	\end{split}
\end{equation}
where $(a)$ is by $-2 \epsilon \times \frac{1}{\epsilon}\langle a,b\rangle = -2 \langle \epsilon a,\frac{1}{\epsilon} b\rangle\leq \epsilon^2\|a\|^2 +\frac{1}{\epsilon^2}\|b\|^2$ and $(b)$ is by Lemma \ref{rho_w}. By taking the telescoping sum, we obtain
\begin{equation}\label{xx_bar_squared}
	\begin{split}
		\| \mathbf{x}_{k+1}-\mathbf{1}\bar{x}_{k+1}\|^2 &\leq \bigg(\frac{1+\rho_w^2}{2}\bigg)^{k+1}\| \mathbf{x}_0 - \mathbf{1}\bar{x}_0\|^2+\frac{2\rho_w^2}{1-\rho_w^2}\sum_{j=0}^{k}\bigg(\frac{1+\rho_w^2}{2}\bigg)^{j+1}\eta_{k-j}^2\| \mathbf{y}_{k-j}-\mathbf{1}\bar{y}_{k-j} \|^2.
	\end{split}
\end{equation}
By the algorithm's updates of $\mathbf{y}_{k+1}$, we can write it in terms of current and all previous gradient estimates as 
\begin{equation}
	\begin{split}
		\mathbf{y}_{k+1} &= W \mathbf{y}_{k}+\mathbf{g}_{k+1}-\mathbf{g}_{k}\\
		&= W (W\mathbf{y}_{k-1}+\mathbf{g}_{k}-\mathbf{g}_{k-1})+\mathbf{g}_{k+1}-\mathbf{g}_{k}\\
		&= W^2 \mathbf{y}_{k-1}-W\mathbf{g}_{k-1}+(W-I)\mathbf{g}_{k}+\mathbf{g}_{k+1}\\
		&= W^2 (W\mathbf{y}_{k-2}+\mathbf{g}_{k-1}-\mathbf{g}_{k-2})-W\mathbf{g}_{k-1}+(W-I)\mathbf{g}_{k}+\mathbf{g}_{k+1}\\
		&= W^3 \mathbf{y}_{k-2}-W^2 \mathbf{g}_{k-2}+ W(W-I)\mathbf{g}_{k-1}+(W-I)\mathbf{g}_{k}+\mathbf{g}_{k+1}\\
		&= \ldots\\
		&= W^{k+1}\mathbf{y}_0 -W^{k}\mathbf{g}_{0}+\sum_{j=0}^{k-1}W^{j}(W-I)\mathbf{g}_{k-j}+\mathbf{g}_{k+1}\\
		&= W^{k}(W-I)\mathbf{g}_0 +\sum_{j=0}^{k-1}W^{j}(W-I)\mathbf{g}_{k-j}+\mathbf{g}_{k+1}\\
		&=\sum_{j=0}^{k}W^{j}(W-I)\mathbf{g}_{k-j}+\mathbf{g}_{k+1},
	\end{split}
\end{equation}
then substituting in $\mathbf{y}_{k}-\mathbf{1}\bar{y}_{k}$, we obtain
\begin{equation}
	\begin{split}
		\mathbf{y}_{k}-\mathbf{1}\bar{y}_{k} &= \sum_{j=0}^{k-1}W^{j}(W-I)\mathbf{g}_{k-1-j}+\mathbf{g}_{k}-\sum_{j=0}^{k-1}\frac{1}{n}\mathbf{1}\mathbf{1}^T W^{j}(W-I)\mathbf{g}_{k-1-j}-\frac{1}{n}\mathbf{1}\mathbf{1}^T\mathbf{g}_{k} \\	
		&= \sum_{j=0}^{k-1}W^{j}(W-I)\mathbf{g}_{k-1-j}+\mathbf{g}_{k}-\sum_{j=0}^{k-1}\frac{1}{n}\mathbf{1}\mathbf{1}^T (W-I)\mathbf{g}_{k-1-j}-\frac{1}{n}\mathbf{1}\mathbf{1}^T\mathbf{g}_{k}\\
		&= \sum_{j=0}^{k-1}(W^{j}-\frac{1}{n}\mathbf{1}\mathbf{1}^T)(W-I)\mathbf{g}_{k-1-j}+\mathbf{g}_{k}-\frac{1}{n}\mathbf{1}\mathbf{1}^T\mathbf{g}_{k}\\
		&= \sum_{j=0}^{k-1}(W-\frac{1}{n}\mathbf{1}\mathbf{1}^T)^{j}(W-I)\mathbf{g}_{k-1-j}+\mathbf{g}_{k}-\mathbf{1}\bar{g}_{k},\\
	\end{split}
\end{equation}
where the last equality is due to the matrix $W$ being doubly stochastic (Assumption \ref{network}) and can be proved by recursion:

$(W-\frac{1}{n}\mathbf{1}\mathbf{1}^T)^{j+1}=(W^{j}-\frac{1}{n}\mathbf{1}\mathbf{1}^T)(W-\frac{1}{n}\mathbf{1}\mathbf{1}^T) = W^{j+1}-\frac{1}{n}W^j \mathbf{1}\mathbf{1}^T-\frac{1}{n}\mathbf{1}\mathbf{1}^T W+\frac{1}{n}\mathbf{1}\mathbf{1}^T = W^{j+1}-\frac{2}{n}\mathbf{1}\mathbf{1}^T+\frac{1}{n}\mathbf{1}\mathbf{1}^T =W^{j+1}-\frac{1}{n}\mathbf{1}\mathbf{1}^T$.

By Lemma \ref{rho_w},
\begin{equation}
	\begin{split}
		\|\mathbf{y}_{k}-\mathbf{1}\bar{y}_{k}\|&\leq \sum_{j=0}^{k-1}\|(W-\frac{1}{n}\mathbf{1}\mathbf{1}^T)^{j}(W-I)\mathbf{g}_{k-1-j}\|+\|\mathbf{g}_{k}-\mathbf{1}\bar{g}_{k}\|\\
		&\leq \sum_{j=0}^{k-1}\rho_w^j\|(W-I)\mathbf{g}_{k-1-j}\|+\|\mathbf{g}_{k}-\mathbf{1}\bar{g}_{k}\|.\\
	\end{split}
\end{equation}
From Lemma \ref{grdt_bnd}, we have $\|\mathbf{g}_{k}\|^2<\infty$ almost surely.
\begin{equation}
	\begin{split}
		\|\mathbf{g}_{k}-\mathbf{1}\bar{g}_{k}\|^2 &= \sum_{i=1}^{n}\|g_{i,k}-\frac{1}{n}\sum_{j=1}^{n}g_{j,k}\|^2 \\
		&= \sum_{i=1}^{n}\bigg(\|g_{i,k}\|^2-2\langle g_{i,k}, \frac{1}{n}\sum_{j=1}^{n}g_{j,k}\rangle+\|\bar{g}_{k}\|^2\bigg)\\
		&= \|\mathbf{g}_{k}\|^2-2n\|\bar{g}_{k}\|^2+n\|\bar{g}_{k}\|^2\\
		&= \|\mathbf{g}_{k}\|^2-n\|\bar{g}_{k}\|^2 \\
		&\leq \|\mathbf{g}_{k}\|^2 \\
		&\leq M'^2<\infty	\\
	\end{split}
\end{equation}
Thus, we get
\begin{equation}\label{y_limit}
	\begin{split}
		\|\mathbf{y}_{k}-\mathbf{1}\bar{y}_{k}\|
		&\leq \frac{M'}{1-\rho_w}\|(W-I)\|+M'\\
		&= G <\infty,
	\end{split}
\end{equation}
where the first term is the sum of a geometric series as $\rho_w<1$. 
\begin{enumerate}
	\item \textbf{Proof of $\lim_{k\rightarrow\infty}\|\mathbf{x}_k-\mathbf{1}\bar{x}_k\|^2 =0 $}
	
	From (\ref{x_fct_previous}), we can write
	\begin{equation}\label{fcts_previous}
		\begin{split}
			\| \mathbf{x}_{k+1}-\mathbf{1}\bar{x}_{k+1}\|^2 
			&\leq \frac{1+\rho_w^2}{2}\| \mathbf{x}_k - \mathbf{1}\bar{x}_k\|^2+\eta_k^2\frac{(1+\rho_w^2)\rho_w^2}{1-\rho_w^2}\| \mathbf{y}_k-\mathbf{1}\bar{y}_k \|^2\\
			\| \mathbf{x}_{k}-\mathbf{1}\bar{x}_{k}\|^2 
			&\leq \frac{1+\rho_w^2}{2}\| \mathbf{x}_{k-1} - \mathbf{1}\bar{x}_{k-1}\|^2+\eta_{k-1}^2\frac{(1+\rho_w^2)\rho_w^2}{1-\rho_w^2}\| \mathbf{y}_{k-1}-\mathbf{1}\bar{y}_{k-1} \|^2\\
			&\ldots\\
			\| \mathbf{x}_{1}-\mathbf{1}\bar{x}_{1}\|^2 
			&\leq \frac{1+\rho_w^2}{2}\| \mathbf{x}_{0} - \mathbf{1}\bar{x}_{0}\|^2+\eta_{0}^2\frac{(1+\rho_w^2)\rho_w^2}{1-\rho_w^2}\| \mathbf{y}_{0}-\mathbf{1}\bar{y}_{0} \|^2.\\
		\end{split}
	\end{equation}
	If we add all inequalities in (\ref{fcts_previous}), we get
	\begin{equation}\label{sum_ineqs}
		\begin{split}
			\| \mathbf{x}_{k+1}-\mathbf{1}\bar{x}_{k+1}\|^2 
			&\leq -\frac{1-\rho_w^2}{2}\sum_{i=1}^{k}\| \mathbf{x}_i - \mathbf{1}\bar{x}_i\|^2+\frac{1+\rho_w^2}{2}\| \mathbf{x}_{0} - \mathbf{1}\bar{x}_{0}\|^2+\frac{(1+\rho_w^2)\rho_w^2}{1-\rho_w^2}\sum_{i=0}^{k}\eta_i^2\| \mathbf{y}_i-\mathbf{1}\bar{y}_i \|^2\\
			&\overset{(a)}{\leq} -\frac{1-\rho_w^2}{2}\sum_{i=1}^{k}\| \mathbf{x}_i - \mathbf{1}\bar{x}_i\|^2+\frac{1+\rho_w^2}{2}\| \mathbf{x}_{0} - \mathbf{1}\bar{x}_{0}\|^2+G^2 \frac{(1+\rho_w^2)\rho_w^2}{1-\rho_w^2}\sum_{i=0}^{k}\eta_i^2\\
		\end{split}
	\end{equation}
	
	where $(a)$ is due to (\ref{y_limit}). Let $k\rightarrow\infty$, then the second and third terms are bounded by Assumption \ref{step_sizes}. We then consider the only two possibilities:  $\sum_i \| \mathbf{x}_{i} - \mathbf{1}\bar{x}_{i}\|^2$ either diverges or converges.
	
	Assume the hypothesis \textit{H}) "$\sum_i \| \mathbf{x}_{i} - \mathbf{1}\bar{x}_{i}\|^2$ diverges." to be true. This implies
	\begin{equation}
		\| \mathbf{x}_{k+1}-\mathbf{1}\bar{x}_{k+1}\|^2 <-\infty,
	\end{equation}
	as $-\frac{1-\rho_w^2}{2}<0$. However, $\| \mathbf{x}_{k+1}-\mathbf{1}\bar{x}_{k+1}\|^2 $ is positive. Thus, hypothesis \textit{H} cannot be true and $\sum_i\| \mathbf{x}_{i} - \mathbf{1}\bar{x}_{i}\|^2$ converges. We conclude that  $\lim_{k\rightarrow\infty}\|\mathbf{x}_k-\mathbf{1}\bar{x}_k\|^2 =0 $ almost surely.
	
	\item \textbf{Proof of $\sum_{k=0}^{\infty}\gamma_{k}\eta_k\| \mathbf{x}_k-\mathbf{1}\bar{x}_k\|^2<\infty$}
	
	Substituting (\ref{xx_bar_squared}) into the sum $\sum_{k=0}^{\infty}\gamma_{k}\eta_k\| \mathbf{x}_k-\mathbf{1}\bar{x}_k\|^2$, we get 
	\begin{equation}\label{bnd_sum_xx}
		\begin{split}
			&\sum_{k=0}^{\infty}\gamma_{k}\eta_k\| \mathbf{x}_k-\mathbf{1}\bar{x}_k\|^2 \\
			= &\gamma_{0}\eta_0\| \mathbf{x}_0-\mathbf{1}\bar{x}_0\|^2+ \sum_{k=1}^{\infty}\gamma_{k}\eta_k\| \mathbf{x}_k-\mathbf{1}\bar{x}_k\|^2 \\
			\leq &\gamma_{0}\eta_0\| \mathbf{x}_0-\mathbf{1}\bar{x}_0\|^2+ \sum_{k=1}^{\infty}\gamma_{k}\eta_k\bigg[\bigg(\frac{1+\rho_w^2}{2}\bigg)^{k}\| 	\mathbf{x}_0 - \mathbf{1}\bar{x}_0\|^2+G^2\frac{2\rho_w^2}{1-\rho_w^2}\sum_{j=0}^{k-1}\bigg(\frac{1+\rho_w^2}{2}\bigg)^{j+1}\eta_{k-1-j}^2\bigg]\\
			\leq &\gamma_{0}\eta_0\| \mathbf{x}_0-\mathbf{1}\bar{x}_0\|^2+\gamma_{0}\eta_0\frac{1+\rho_w^2}{1-\rho_w^2}\|\mathbf{x}_0 - \mathbf{1}\bar{x}_0\|^2 +G^2\frac{2\rho_w^2}{1-\rho_w^2}\sum_{k=1}^{\infty}\gamma_{k}\eta_k\sum_{j=0}^{k-1}\bigg(\frac{1+\rho_w^2}{2}\bigg)^{j+1}\eta_{k-1-j}^2\\
			= &\frac{2\gamma_{0}\eta_0}{1-\rho_w^2}\|\mathbf{x}_0 - \mathbf{1}\bar{x}_0\|^2 +G^2\frac{2\rho_w^2}{1-\rho_w^2}\sum_{k=1}^{\infty}\gamma_{k}\eta_k\sum_{j=0}^{k-1}\bigg(\frac{1+\rho_w^2}{2}\bigg)^{j+1}\eta_{k-1-j}^2,
		\end{split}
	\end{equation}
	where the last inequality is due to the fact that $\gamma_k$ and $\eta_k$ are both decreasing step-sizes and we have a sum of a geometric series of ratio $\frac{1+\rho^2}{2}<1$. We then study the last term,
	\begin{equation}\label{bnd_sum_xx2}
		\begin{split}
			\sum_{k=1}^{\infty}\gamma_{k}\eta_k\sum_{j=0}^{k-1}\bigg(\frac{1+\rho_w^2}{2}\bigg)^{j+1}\eta_{k-1-j}^2
			\leq &\sum_{k=1}^{\infty}\gamma_{k}\sum_{j=0}^{k-1}\bigg(\frac{1+\rho_w^2}{2}\bigg)^{j+1}\eta_{k-1-j}^3\\
			= &\sum_{k=1}^{\infty}\gamma_{k}\sum_{j=1}^{k}\bigg(\frac{1+\rho_w^2}{2}\bigg)^{k-j+1}\eta_{j-1}^3\\
			=
			&\sum_{j=1}^{\infty}\eta_{j-1}^3\sum_{k=j}^{\infty}\gamma_{k}\bigg(\frac{1+\rho_w^2}{2}\bigg)^{k-j+1}\\
			\leq
			&\gamma_{0}\sum_{j=1}^{\infty}\eta_{j-1}^3\sum_{k=j}^{\infty}\bigg(\frac{1+\rho_w^2}{2}\bigg)^{k-j+1}\\
			=
			&\gamma_{0}\frac{1+\rho_w^2}{1-\rho_w^2}\sum_{j=1}^{\infty}\eta_{j-1}^3\\
			< &\infty,
		\end{split}
	\end{equation}
	as $\sum\eta_k^2$ converges by Assumption \ref{step_sizes}.
	
	Finally, $\sum_{k=0}^{\infty}\gamma_{k}\eta_k\| \mathbf{x}_k-\mathbf{1}\bar{x}_k\|^2<\infty$.
\end{enumerate}
	
\end{document}